\documentclass[11pt]{article}

\usepackage[margin=1in]{geometry}

\usepackage[utf8]{inputenc}
\usepackage[T1]{fontenc}
\usepackage{amsmath,amssymb,amsfonts,amsthm}
\newtheorem{proposition}{Proposition}
\newtheorem{remark}{Remark}
\newtheorem{lemma}{Lemma}
\newtheorem{corollary}{Corollary}
\newtheorem{assumption}{Assumption}
\usepackage{graphicx}
\usepackage{booktabs}
\usepackage{multirow}
\usepackage{hyperref}
\usepackage{cleveref}
\usepackage{xcolor}
\usepackage{algorithm}
\usepackage{algorithmic}
\usepackage{caption}
\usepackage{subcaption}
\usepackage{enumitem}
\usepackage{microtype}
\usepackage{float}
\usepackage{tikz}
\usepackage{natbib}

\graphicspath{{plots/}}

\newcommand{\ours}{PEARL}
\newcommand{\xu}{\mathbf{x}_u}
\newcommand{\zk}{\mathbf{z}_k}
\newcommand{\rhat}{\hat{r}}

\title{PEARL: Prediction Error-Adaptive Regularized Learning for Selective Forgetting under Partial Preference Drift}

\author{
  Rashmi Gangadharaiah \\
  AWS \\
  \texttt{rgangad@amazon.com}
  \and
  Balakrishnan Narayanaswamy \\
  AWS \\
  \texttt{muralibn@amazon.com}
}

\begin{document}

\maketitle

\begin{abstract}
User preferences are not static: they shift as needs, context, and tastes evolve. This is particularly important in personalized recommendation, appropriate LLM response generation, and RLHF-trained reward models, all of which can become miscalibrated as user preferences drift, requiring online adaptation. Since only the selected response receives feedback, the system must balance exploiting its current best against exploring alternatives. Existing non-stationary bandit methods forget uniformly across all model dimensions, but real preference shifts are typically partial: some dimensions go stale while others remain valid. Neural contextual bandits accumulate confidence monotonically via the Gram matrix over learned features, with no mechanism to direct uncertainty to where predictions have actually gone wrong. We propose \ours{} (\textbf{P}rediction \textbf{E}rror-\textbf{A}daptive \textbf{R}egularized \textbf{L}earning), which reverses the Gram update along feature directions where predictions are surprisingly wrong, while leaving confidence intact everywhere else. We prove that (i) any non-adaptive policy suffers linear regret after a preference shift, (ii) no fixed exploration rate achieves low regret in both stationary and non-stationary regimes simultaneously, and (iii) \ours{}'s cumulative regret scales with the number of shifted users $N_s$, not the total population $N$, with a one-interaction detection delay. On PRISM, a personalized LLM response selection benchmark (1,500 users, 21 LLMs), with moderate preference drift, \ours{} beats greedy selection by $+$4.0 points and the best neural bandit baseline by $+$2.5 points ($p{<}0.001$), while performing on par with greedy when preferences are stable. It beats sliding-window and discounted learning approaches by 3 to 10 points, showing that selective forgetting outperforms forgetting uniformly across all model dimensions. The selection policy is a ${\sim}$100K-parameter MLP that runs in milliseconds on a CPU, leaves LLM weights frozen, and works with any API-based model.
\end{abstract}

\section{Introduction}
\label{sec:intro}
Neural contextual bandits learn which actions to take by maintaining a Gram matrix over learned features~\citep{zhou2020neural}. Each observation tightens the matrix, reducing epistemic uncertainty in the corresponding feature direction, which is the right behavior when user preferences are stable. But the update is irreversible. Once the model has seen enough data along a feature direction, the exploration bonus in that direction stays small permanently. If the user's preferences then change in precisely that direction, the model will keep exploiting a now-wrong prediction and the Gram matrix offers no mechanism to course-correct.

\begin{figure}[!ht]
\centering
\begin{subfigure}[t]{0.52\textwidth}
    \centering
    \includegraphics[width=\textwidth]{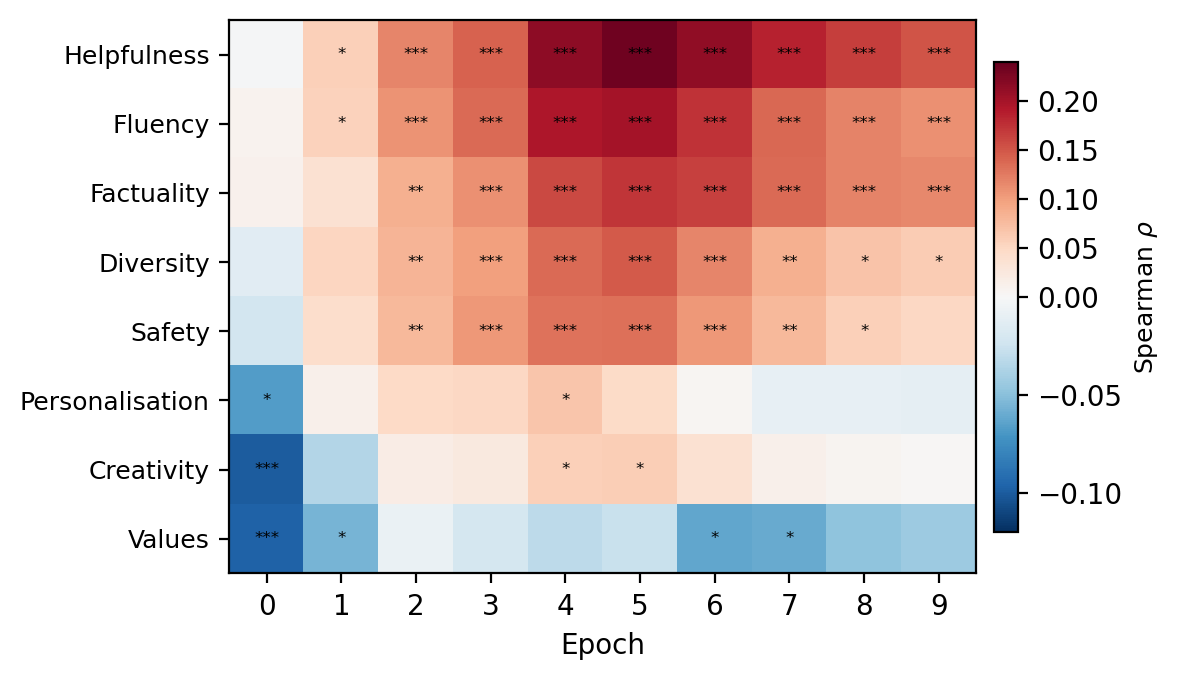}
    \caption{Spearman $\rho$ between preference strength and prediction error for shifted users (20 seeds). Helpfulness, fluency, and factuality show persistent correlations; creativity and values do not.}
    \label{fig:pref_heatmap}
\end{subfigure}
\hfill
\begin{subfigure}[t]{0.44\textwidth}
    \centering
    \includegraphics[width=\textwidth]{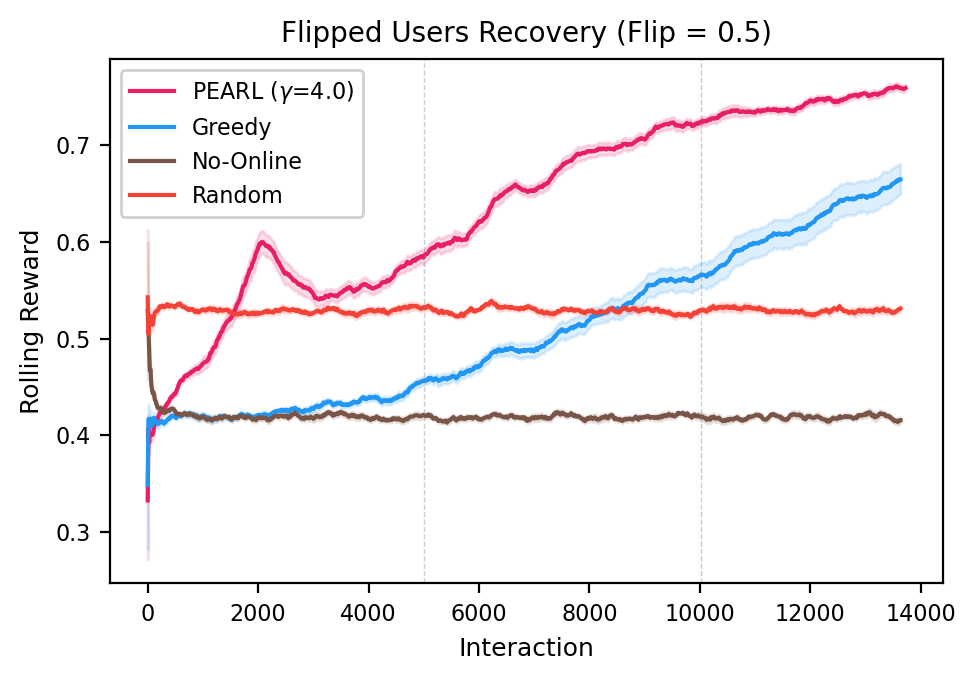}
    \caption{Recovery trajectory for flipped users (flip 0.5, 20 seeds, rolling window of 500). \ours{} separates from Greedy by interaction 2,000, recovering faster where interactions are costly. No-Online flatlines.}
    \label{fig:recovery}
\end{subfigure}
\caption{\textbf{(a)}~Real preference data exhibits partial stationarity: dimensions go stale at different rates. \textbf{(b)}~\ours{} exploits this structure, recovering faster than Greedy on shifted users.}
\label{fig:overview}
\end{figure}

\Cref{fig:pref_heatmap} illustrates this failure concretely. For users whose preferences shift, we measure the correlation between how much a user cares about each preference dimension and the model's prediction error for that user. On dimensions the model had learned well before the shift (Helpfulness, Factuality, Fluency), prediction error is highest for the users who care most, and this relationship persists as the model continues to train. The model is most wrong precisely where it matters most to each user, and it remains wrong because the Gram matrix is most confident in those directions and has stopped exploring them. Other dimensions (Creativity, Values) show no such pattern. The damage is concentrated where confidence is highest, yet the standard Gram update cannot distinguish stale confidence from valid.

Existing non-stationary bandit methods apply uniform forgetting: sliding windows~\citep{garivier2011upper}, discounted Gram matrices~\citep{russac2019weighted}, or periodic restarts. When \emph{every} dimension may change, this is reasonable. But preference drift is rarely so uniform.

A user who starts wanting concise answers and later prefers detailed explanations has not changed their preference for factual accuracy. A moviegoer whose genre tastes shift from action to romance still cares about compelling storytelling. In practice, some dimensions of the reward function go stale while others remain perfectly valid. We call this \emph{partial stationarity}. The stakes are highest when preferences are deeply personal: in the PRISM dataset~\citep{kirk2024prism}, the same LLM response can score 85 from one user and 40 from another across 1,500 participants, and the bottleneck is not generation but selection. Under partial stationarity, uniform forgetting discards valid information alongside stale, forcing re-exploration of dimensions that never changed. The cost scales with total dimensionality, not the number of shifted dimensions (\Cref{prop:forgetting_efficiency}, \Cref{app:directional_efficiency}). This raises a concrete question: \emph{can a neural bandit selectively forget what has gone stale while preserving what remains valid?} To our knowledge, no prior work formalizes partial stationarity as a distinct setting or designs a forgetting mechanism tailored to it.

We propose \ours{} (\textbf{P}rediction \textbf{E}rror-\textbf{A}daptive \textbf{R}egularized \textbf{L}earning), a neural contextual bandit that breaks the one-way accumulation in the Gram matrix. After each interaction, if the prediction error exceeds the user's running average, \ours{} injects uncertainty back along the feature direction where the error occurred. Confidence in all other directions is untouched. An error-adaptive acquisition function complements this: per-user prediction errors modulate the exploration bonus, so well-predicted users get near-greedy selection while poorly-predicted users explore aggressively. When preferences are stable, both mechanisms go dormant.

We evaluate on two domains under controlled preference drift: LLM response selection (PRISM; 1,500 users, 21 LLMs) and a simulation with known ground-truth preferences. On PRISM at moderate drift, \ours{} beats greedy by $+$4.0 points and NeuralUCB by $+$2.5 ($p{<}0.001$; \Cref{tab:flip}); static models without online updates fall below random (0.480 vs.\ 0.537). The regret decomposition predicts a $d/s$ advantage when only $s$ of $d$ dimensions shift (\Cref{prop:forgetting_efficiency}, \Cref{app:directional_efficiency}); the measured 3--10 point gap confirms this.

\noindent\textbf{Contributions.}
\begin{enumerate}[leftmargin=*,itemsep=1pt,topsep=2pt,parsep=0pt]
    \item The \textbf{bidirectional Gram update}: a mechanism that injects uncertainty back into the Gram matrix along feature directions where prediction error is surprisingly high, while preserving confidence in well-calibrated directions (\Cref{sec:bidir_gram}). Setting the injection rate to zero recovers standard NeuralUCB.

    \item \textbf{Error-adaptive exploration}: a per-user prediction error tracker that modulates the UCB bonus, achieving near-greedy behavior under stationarity and rapid recovery under drift (\Cref{sec:bandit}).

    \item Evaluation on \textbf{two domains} (LLM personalization and controlled simulation) with a systematic ablation chain showing that \ours{} dominates all baselines under partial stationarity and incurs minimal cost under stationarity (\Cref{sec:results}).

    \item \textbf{Practical deployability}: a ${\sim}$100K-parameter MLP that operates on precomputed features, requiring no GPU and no modification to the underlying models. For LLM personalization, a small reward model learns user preferences online, requiring no expensive LLM updates when those preferences change.
\end{enumerate}


\section{Related Work}
\label{sec:related}
LinUCB~\citep{li2010contextual} introduced contextual bandits with linear reward models and confidence-based exploration. NeuralUCB~\citep{zhou2020neural} and NeuralTS~\citep{zhang2020neural} extend this to neural networks using last-layer Gram-matrix uncertainty, but assume stationarity. SW-UCB~\citep{garivier2011upper} handles non-stationarity via sliding windows; \citet{russac2019weighted} use discounted Gram matrices in \emph{linear} bandits. Both forget indiscriminately across all dimensions. \ours{} draws on classical directional forgetting in RLS~\citep{kulhavy1984tracking,bittanti1990convergence}, developed for tracking slowly varying parameters in linear system identification, but extends it to neural features with error-conditional triggering and a spectral cap that create a self-correcting loop. Unlike continual learning~\citep{kirkpatrick2017overcoming,zenke2017continual}, which protects important \emph{weights} across tasks, our update operates at the \emph{uncertainty} level, restoring epistemic confidence in stale directions while leaving weights to be updated by gradient descent. Time-varying matrix factorization~\citep{koren2009collaborative}, contextual signals~\citep{hariri2015adapting}, and recurrent bandit models~\citep{wu2018learning} operate at the global level and cannot identify which feature dimensions have gone stale.

Best-of-N sampling~\citep{stiennon2020learning} uses no exploration signal. URM~\citep{lou2024uncertainty} adds heteroscedastic heads, \citet{zhai2023uncertainty} propose LoRA ensembles, and deep ensembles~\citep{lakshminarayanan2017simple} provide well-calibrated uncertainty~\citep{ovadia2019can} but scale linearly with ensemble size. None were designed for bandits; we adapt them as baselines in \Cref{sec:baselines}. \ours{} uses single-network Gram-matrix uncertainty, avoiding ensemble overhead while enabling directional forgetting.

RLHF~\citep{ouyang2022training,bai2022training} and DPO~\citep{rafailov2023direct} learn a single policy from aggregated preferences. Offline personalization methods (MRM via meta-learning, VRF via variational bases) learn user-specific rewards but assume static preferences. VPL and SDPO fine-tune the LLM itself, which is expensive and impractical for API-only access. MiCRo~\citep{shen2025micro} and T-POP~\citep{qu2025t} maintain frozen-model reward models but lack per-user epistemic uncertainty or non-stationarity handling. Pluralistic alignment~\citep{sorensen2024roadmap,feng2024modular,wang2026personalized} targets group-level rather than individual preferences.

\section{Method}
\label{sec:method}
\subsection{Problem Setup}
\label{sec:problem}

Consider a contextual bandit with context $\xu \in \mathbb{R}^{d_u}$ (e.g., user features) and $K$ arms with feature vectors $\zk \in \mathbb{R}^{d_z}$ (e.g., candidate items or responses). At each round, the learner selects an arm $k^*$ and observes only the reward $r(k^*) \in [0,1]$ for that arm. The goal is to maximize cumulative reward. We focus on the \emph{partially stationary} setting where user preferences may change over time, so that a model trained on historical data can become miscalibrated.

\subsection{Reward Model with Gram-Matrix Uncertainty}
\label{sec:reward_model}
We model the reward function with a single MLP $f_\theta$, parameterized by $\theta$, mapping concatenated context and arm features to a $d$-dimensional representation, followed by a linear head $h: \mathbb{R}^d \to \mathbb{R}$:
\begin{equation}
    g_k = f_\theta([\xu; \zk]) \in \mathbb{R}^d, \qquad \rhat(\xu, \zk; \theta) = \text{sigmoid}(h(g_k))
    \label{eq:backbone}
\end{equation}
We write $\mu_k = \rhat(\xu, \zk; \theta)$ for brevity in subsequent equations. While we use a two-layer MLP ($d{=}128$, dropout 0.1), the framework is agnostic to the reward architecture. We compare against heteroscedastic (URM), meta-learned (MRM), and variational (VRF) reward heads in \Cref{sec:baselines}.

\textbf{Gram-matrix uncertainty.} Following NeuralUCB, we maintain a Gram matrix $A$ over the network's last-layer features to quantify epistemic uncertainty:
\begin{equation}
    A = \lambda I + \sum_{t=1}^{T} g_t g_t^\top, \qquad
    \sigma_k = \sqrt{g_k^\top A^{-1} g_k}
    \label{eq:gram}
\end{equation}
where $T$ is the number of observations so far and $\lambda$ is a regularization parameter. The uncertainty $\sigma_k$ is small in feature-space directions where many observations have been collected, and large in unexplored directions. This captures \emph{epistemic} uncertainty, the appropriate signal for exploration. We maintain $A^{-1}$ directly via rank-1 Sherman-Morrison updates (\Cref{sec:bidir_gram}).

\textbf{Context representation.} We construct a hybrid context vector combining observed features with learned latent factors:
\begin{equation}
    \xu = \left[\mathbf{s}_u \,;\; e_\phi(u)\right] \in \mathbb{R}^{d_u}
    \label{eq:user_repr}
\end{equation}
where $\mathbf{s}_u$ is an observed feature vector and $e_\phi$ is a learned embedding trained jointly with the reward model. Embedding dampening $(1 - \varepsilon_u) \cdot e_\phi(u)$ down-weights learned embeddings when the model is uncertain, falling back to observed features. Dataset-specific dimensions are given in \Cref{sec:setup}.

\textbf{Training.} We pretrain on historical data using Bradley-Terry pairwise loss~\citep{bradley1952rank}, mirroring the reward model training stage in RLHF, which is natural when multiple ratings per context are available. During the online phase, where the model is deployed and learns continuously from user interactions, only the reward $r$ for the selected arm is observed, so pairwise comparisons are no longer available and we switch to pointwise MSE over the online dataset $\mathcal{D}_{\text{on}}$:
\begin{equation}
    \mathcal{L}_{\text{online}}(\theta) = \frac{1}{|\mathcal{D}_{\text{on}}|}\sum_{(\xu, \zk, r) \in \mathcal{D}_{\text{on}}} \left(\rhat(\xu, \zk; \theta) - r\right)^2
    \label{eq:mse_loss}
\end{equation}

\subsection{Error-Adaptive Exploration (\ours{})}
\label{sec:bandit}

Standard UCB explores uniformly across all users, wasting budget on users whose rewards are already well-modeled. \ours{} instead monitors its own prediction errors per user and scales exploration intensity accordingly.

\textbf{Per-user drift detector.} For each user $u$, we maintain an exponential moving average (EMA) of absolute prediction errors:
\begin{equation}
    \varepsilon_u^{(t)} = \beta \cdot \varepsilon_u^{(t-1)} + (1 - \beta) \cdot \ell_u^{(t)}, \qquad \ell_u^{(t)} = \left|r_{k^*}^{(t)} - \mu_{k^*}^{(t)}\right|
    \label{eq:ema_error}
\end{equation}
where $\beta{=}0.3$ is the smoothing factor (robust to choice; \Cref{app:beta_ablation}), initialized at $\varepsilon_u^{(0)} = 0.5$ (maximum uncertainty on $[0,1]$ rewards, encouraging exploration before any observations). When predictions are accurate, $\varepsilon_u$ decays toward zero; when user preferences shift, prediction errors spike, providing an automatic drift signal.

\textbf{\ours{} acquisition function.} Arm selection maximizes an error-adaptive acquisition function:
\begin{equation}
    k^* = \arg\max_{k \in \{1,\ldots,K\}} \; \underbrace{\mu_k(\xu, \zk)}_{\text{exploitation}} \;+\; \underbrace{\gamma \cdot \varepsilon_u^{(t)} \cdot \sigma_k(\xu, \zk)}_{\text{error-adaptive exploration bonus}}
    \label{eq:acquisition}
\end{equation}
The bonus factors into $\gamma$ (global sensitivity), $\varepsilon_u^{(t)}$ (per-user prediction quality), and $\sigma_k$ (per-arm epistemic uncertainty). When the model predicts well ($\varepsilon_u \to 0$), selection reduces to greedy; when predictions are poor, the bonus upweights uncertain arms.

\textbf{Self-correcting dynamics.}
The mechanism is self-correcting: when rewards shift, errors spike, $\varepsilon_u$ rises, and exploration increases; as new data arrives, $\varepsilon_u$ falls and selection reverts to near-greedy. Under stationarity, $\varepsilon_u \to 0$ and exploration adds no cost. The tracker detects shifts within a single interaction at $\beta{=}0.3$ (\Cref{lem:response_time}, \Cref{app:response_time}). A per-round regret decomposition (\Cref{rem:regret}, \Cref{app:bayesian_regret}) yields an adaptive bound (\Cref{prop:adaptive_regret}, \Cref{app:adaptive_regret}) where cumulative regret scales with the number of \emph{shifted} users $N_s$, not the total $N$.

\subsection{Bidirectional Gram Update}
\label{sec:bidir_gram}

\textbf{Intuition: two dimensions, one shifts.} Suppose a user cares about two attributes, verbosity and formality, and the Gram matrix has accumulated confidence in both directions. Now the user's verbosity preference flips. The model is wrong along verbosity but still correct along formality. Uniform forgetting (sliding windows, discounted Gram) would reduce confidence in \emph{both} directions, forcing the model to re-learn formality from scratch even though it was never wrong there. The wasted re-exploration scales with $d$, not the number of shifted dimensions $s$ (\Cref{prop:forgetting_efficiency}, \Cref{app:directional_efficiency}). What we want: grow uncertainty back along verbosity and leave formality alone.

In standard NeuralUCB, the Gram matrix can only accumulate observations: each new data point \emph{shrinks} uncertainty via the Sherman-Morrison rank-1 update:
\begin{equation}
    A^{-1} \leftarrow A^{-1} - \frac{(A^{-1} g_k)(g_k^\top A^{-1})}{1 + g_k^\top A^{-1} g_k}
    \label{eq:sherman_morrison}
\end{equation}
Once the model has seen data in a feature direction, $\sigma_k$ stays permanently low there, even if a preference shift makes the model wrong in exactly that direction. The EMA tracker $\varepsilon_u$ compensates by scaling \emph{all} directions equally, but this uniform scalar adjustment cannot distinguish directions where the model is wrong from those where it remains correct.

The \textbf{bidirectional Gram update} fixes this by allowing uncertainty to \emph{grow back} in directions where the model is wrong. After each interaction, we perform the standard shrink (\cref{eq:sherman_morrison}), then conditionally inject uncertainty:
\begin{equation}
    \text{If } |\rhat_k - r_k| > \varepsilon_u \text{ and } \varepsilon_u > 0.05: \quad
    A^{-1} \leftarrow A^{-1} + \eta \cdot (r_k - \rhat_k)^2 \cdot g_k g_k^\top
    \label{eq:injection}
\end{equation}
where $\eta$ is the injection rate. The injection fires only when the prediction error exceeds the context's running average, i.e., when the model is \emph{surprisingly} wrong. Three properties make this work. First, it is \textbf{directional}: uncertainty grows only along $g_k$, the feature direction where the error occurred. Second, it is \textbf{error-scaled}: larger errors produce proportionally larger injections ($\propto (r_k - \rhat_k)^2$). Third, it is \textbf{self-correcting}: as exploration improves predictions, errors shrink, injection stops triggering, and $\sigma_k$ decays back through normal Sherman-Morrison updates.

\textbf{Spectral safeguard.} Repeated injections could cause $A^{-1}$ eigenvalues to grow unboundedly. We apply a spectral cap after each injection:
\begin{equation}
    A^{-1} \leftarrow U \, \text{diag}\!\left(\min(\lambda_i, 1/\lambda)\right) U^\top
    \label{eq:spectral_cap}
\end{equation}
where $U \Lambda U^\top = A^{-1}$ is the eigendecomposition and $1/\lambda$ is the prior uncertainty. This ensures that injected uncertainty never exceeds the prior: the model can ``forget'' what it learned in a direction, but cannot become more uncertain than before seeing any data. Since $\eta (r_k - \rhat_k)^2 g_k g_k^\top$ is positive semidefinite, $A^{-1}$ remains positive definite throughout. The eigendecomposition is $O(d^3)$ but with $d{=}128$ it takes under 1ms on CPU; moreover, the cap is applied only when injection fires (${\sim}$40\% of interactions), so the amortized overhead is negligible.

Although $A^{-1}$ is shared across all users, the injection direction $g_k = f_\theta([\xu; \zk])$ is conditioned on the per-user context $\xu$ (\cref{eq:user_repr}), so two users with different preferences produce different $g_k$ vectors even for the same arm; injection for one user therefore does not distort uncertainty estimates for dissimilar users. Classical directional forgetting in RLS injects uncertainty on a fixed schedule over linear features; the bidirectional Gram update differs by operating on nonlinear neural features, firing only when error exceeds the running average, and scaling injection with squared error. Setting $\eta{=}0$ recovers standard NeuralUCB. The full procedure is given in \Cref{alg:bandit}.

\begin{algorithm}[t]
\caption{\ours{}: Online Selection with Bidirectional Gram}
\label{alg:bandit}
\begin{algorithmic}[1]
\REQUIRE Network $f_\theta$, head $h$, context features $\{\xu\}$, interaction stream
\STATE $\mathcal{D} \leftarrow \emptyset$; \; $\varepsilon_u \leftarrow 0.5 \;\forall u$; \; $A^{-1} \leftarrow \frac{1}{\lambda}I$
\FOR{each interaction $(u, \{(k, \zk)\}_{k=1}^K)$}
    \STATE $g_k \leftarrow f_\theta([\xu; \zk])$; \; $\mu_k \leftarrow h(g_k)$; \; $\sigma_k \leftarrow \sqrt{g_k^\top A^{-1} g_k}$ \;\; $\forall k$
    \STATE Select $k^* \leftarrow \arg\max_k [\mu_k + \gamma \cdot \varepsilon_u \cdot \sigma_k]$; observe $r_{k^*}$
    \STATE $A^{-1} \leftarrow A^{-1} - \frac{(A^{-1} g_{k^*})(g_{k^*}^\top A^{-1})}{1 + g_{k^*}^\top A^{-1} g_{k^*}}$ \COMMENT{Shrink}
    \IF{$|r_{k^*} - \mu_{k^*}| > \varepsilon_u$ \AND $\varepsilon_u > 0.05$}
        \STATE $A^{-1} \leftarrow A^{-1} + \eta (r_{k^*} - \mu_{k^*})^2 g_{k^*} g_{k^*}^\top$; spectral cap to $\leq 1/\lambda$ \COMMENT{Grow}
    \ENDIF
    \STATE $\varepsilon_u \leftarrow \beta \varepsilon_u + (1 - \beta)|r_{k^*} - \mu_{k^*}|$; \; add $(\xu, \mathbf{z}_{k^*}, r_{k^*})$ to $\mathcal{D}$
    \STATE Periodically: update $f_\theta, h$ on recent $\mathcal{D}$ (\cref{eq:mse_loss}); rebuild $A^{-1}$
\ENDFOR
\end{algorithmic}
\end{algorithm}

Between network updates, $A^{-1}$ evolves via per-interaction rank-1 updates at $O(d^2)$ cost ($d{=}128$, negligible); network updates retrain on recent observations and rebuild $A^{-1}$ from updated features.

\section{Error Tracker Recovery Analysis}
\label{app:response_time}

\begin{lemma}[Error Tracker Response Time]
\label{lem:response_time}
Suppose user $u$'s preferences shift at time $t_0$, causing prediction errors $\ell_u^{(t)} \geq \delta_{\text{post}}$ for $t \geq t_0$. Then the EMA error tracker (\cref{eq:ema_error}) satisfies:
\begin{equation}
    \varepsilon_u^{(t_0 + \Delta)} \geq \delta_{\text{post}} \cdot (1 - \beta^{\Delta})
    \label{eq:recovery_rate}
\end{equation}
In particular, within $\Delta^* = \lceil \log(0.5) / \log(\beta) \rceil$ interactions, $\varepsilon_u$ reaches at least $\delta_{\text{post}}/2$.
\end{lemma}
\begin{proof}
By induction on \cref{eq:ema_error}: $\varepsilon_u^{(t_0 + \Delta)} \geq (1 - \beta) \cdot \delta_{\text{post}} \cdot \sum_{s=0}^{\Delta-1} \beta^s = \delta_{\text{post}} \cdot (1 - \beta^{\Delta})$. Setting $(1 - \beta^{\Delta}) \geq 1/2$ gives $\Delta^* = \lceil \log(0.5) / \log(\beta) \rceil$. For $\beta{=}0.3$, $\Delta^*{=}1$ (immediate response).
\end{proof}

As the model retrains on exploratory observations, calibration improves and $\varepsilon_u$ decays, automatically tapering exploration. This yields a characteristic \emph{explore-then-exploit recovery} profile: the exploration bonus peaks immediately after the shift and decays as the model adapts. Fixed UCB, by contrast, explores at a rate determined solely by the time-decay schedule $\alpha_t$, which is agnostic to whether a preference shift has occurred.

\section{EMA Smoothing Factor Ablation}
\label{app:beta_ablation}

We ablate the EMA smoothing factor $\beta$ in \cref{eq:ema_error} across $\{0.1, 0.3, 0.5, 0.7\}$ under flip 0.5 (20 seeds, $\gamma{=}4.0$). Results: $\beta{=}0.1$: 0.619, $\beta{=}0.3$: 0.618, $\beta{=}0.5$: 0.618, $\beta{=}0.7$: 0.617. The method is insensitive to $\beta$ within this range (all within 0.2 points), confirming that the error-adaptive mechanism is robust to this hyperparameter. We use $\beta{=}0.3$ throughout, which provides a fast response time ($\Delta^*{=}1$, \cref{lem:response_time}).

\section{Connection to Bayesian Regret Bounds}
\label{app:bayesian_regret}

We prove the regret decomposition underlying \cref{eq:acquisition} (stated informally in \cref{sec:bandit}), then formalize the relationship between \ours{} and classical Bayesian bandit algorithms.

\begin{proposition}[Per-Round Regret Decomposition]
\label{rem:regret}
Under the \ours{} acquisition function (\cref{eq:acquisition}), the per-round regret decomposes as:
$r_u(k^*_{\text{opt}}) - r_u(k^*_t) \leq 2\sup_k|r_u(k) - \mu_k| + \gamma\varepsilon_u^{(t)}\sigma_{k^*_t}$.
\end{proposition}

\begin{proof}
By the selection rule, $\mu_{k^*_t} + \gamma\varepsilon_u^{(t)}\sigma_{k^*_t} \geq \mu_{k^*_{\text{opt}}} + \gamma\varepsilon_u^{(t)}\sigma_{k^*_{\text{opt}}}$, so $\mu_{k^*_{\text{opt}}} - \mu_{k^*_t} \leq \gamma\varepsilon_u^{(t)}\sigma_{k^*_t}$. Decomposing:
\begin{align*}
    r_u(k^*_{\text{opt}}) - r_u(k^*_t) &= \bigl[r_u(k^*_{\text{opt}}) - \mu_{k^*_{\text{opt}}}\bigr] + \bigl[\mu_{k^*_{\text{opt}}} - \mu_{k^*_t}\bigr] + \bigl[\mu_{k^*_t} - r_u(k^*_t)\bigr] \\
    &\leq 2\sup_k|r_u(k) - \mu_k| + \gamma\varepsilon_u^{(t)}\sigma_{k^*_t} \qedhere
\end{align*}
\end{proof}

\paragraph{BayesUCB connection.} We show that our acquisition function (\cref{eq:acquisition}) can be viewed as an instance of BayesUCB~\citep{kaufmann2012bayesian} with a user-adaptive confidence parameter.

\paragraph{BayesUCB formulation.} BayesUCB selects arms by maximizing the $(1-1/t)$-quantile of the posterior predictive distribution:
\begin{equation}
    k^*_{\text{BayesUCB}} = \arg\max_k \; Q_{1-1/t}\!\left(r \mid \xu, \zk, \mathcal{D}\right)
    \label{eq:bayesucb}
\end{equation}
Under a Gaussian approximation to the posterior predictive (justified by the NTK-regime analysis of NeuralUCB~\citep{zhou2020neural}), this reduces to:
\begin{equation}
    k^*_{\text{BayesUCB}} \approx \arg\max_k \left[\mu_k + \Phi^{-1}\!\left(1 - \tfrac{1}{t}\right) \cdot \sigma_k\right]
    \label{eq:bayesucb_gaussian}
\end{equation}
where $\Phi^{-1}$ is the standard normal quantile function. The exploration weight $\Phi^{-1}(1 - 1/t)$ grows as $O(\sqrt{\log t})$, matching the classical UCB1 rate~\citep{auer2002finite}.

\paragraph{\ours{} as user-adaptive BayesUCB with directional forgetting.} Comparing \cref{eq:bayesucb_gaussian} with the \ours{} acquisition function (\cref{eq:acquisition}), we can identify:
\begin{equation}
    \underbrace{\gamma \cdot \varepsilon_u^{(t)}}_{\text{\ours{} exploration weight}} \quad \longleftrightarrow \quad \underbrace{\Phi^{-1}\!\left(1 - \tfrac{1}{t}\right)}_{\text{BayesUCB exploration weight}}
    \label{eq:correspondence}
\end{equation}
The key difference is the source of exploration modulation. BayesUCB uses a \emph{global} time-dependent schedule that decreases uniformly for all users. \ours{} replaces this with a \emph{per-user} data-dependent signal $\varepsilon_u^{(t)}$ that adapts to local prediction quality, combined with a bidirectional Gram update that modulates $\sigma_k$ itself directionally. This has three concrete advantages:

\begin{enumerate}[leftmargin=*]
    \item \textbf{Heterogeneous exploration rates.} In a population with diverse learnability (some users have simple, consistent preferences; others have complex or shifting ones), BayesUCB applies the same confidence level to all users at time $t$. \ours{} automatically allocates more exploration to hard-to-predict users ($\varepsilon_u$ large) and less to easy users ($\varepsilon_u$ small), improving the population-level exploration budget allocation.

    \item \textbf{Partial-stationarity responsiveness.} The BayesUCB quantile $\Phi^{-1}(1-1/t)$ is monotonically increasing in $t$ and cannot respond to preference shifts. In contrast, $\varepsilon_u^{(t)}$ is non-monotone: it can increase when prediction errors spike after a preference change (\cref{eq:recovery_rate}), re-initiating exploration precisely when it is needed.

    \item \textbf{Directional uncertainty recovery.} In standard NeuralUCB, $\sigma_k$ can only decrease as data accumulates. \ours{}'s bidirectional Gram update (\cref{eq:injection}) allows $\sigma_k$ to grow back in feature directions where the model is wrong, providing a second, directional channel for partial-stationarity response that complements the scalar $\varepsilon_u$ modulation. The spectral cap (\cref{eq:spectral_cap}) ensures the injected uncertainty never exceeds the prior, preserving the validity of the confidence bound in Assumption~\ref{asm:calibrated}.
\end{enumerate}

\paragraph{Bayesian cumulative regret.} For a standard $K$-armed contextual bandit with sub-Gaussian rewards, BayesUCB achieves Bayesian cumulative regret~\citep{kaufmann2012bayesian,russo2014learning}:
\begin{equation}
    \text{BayesRegret}(T) = \mathbb{E}_{\theta \sim \pi}\!\left[\sum_{t=1}^{T} r(k^*_{\text{opt}}) - r(k^*_t)\right] = O\!\left(\sqrt{KT \log T}\right)
    \label{eq:bayes_regret}
\end{equation}
where the expectation is over the prior $\pi$ on the reward parameters. Under stationarity, \ours{} inherits a comparable rate, since $\varepsilon_u^{(t)} \to 0$ at a geometric rate (\cref{eq:ema_error}), ensuring that the cumulative exploration cost $\sum_t \gamma \varepsilon_u^{(t)} \sigma_{k^*_t}$ converges. The bidirectional Gram update (\cref{sec:bidir_gram}) preserves this rate under stationarity because the injection condition ($|\rhat_k - r_k| > \varepsilon_u$) is rarely triggered when the model is well-calibrated, and the spectral cap (\cref{eq:spectral_cap}) ensures $\sigma_k$ never exceeds its prior value.

\begin{corollary}[Partially-Stationary Regret Decomposition]
\label{cor:nonstationary_regret}
Under Assumptions~\ref{asm:subgaussian}--\ref{asm:variation} (stated in \cref{app:adaptive_regret}), suppose user preferences undergo $S$ discrete shifts, each inducing a recovery period of at most $\Delta_{\mathrm{rec}}$ interactions. Let $\delta_{\mathrm{post}}$ denote the maximum post-shift calibration error. Then the cumulative regret of \ours{} decomposes as:
\begin{equation}
    \mathrm{Regret}_{\ours{}}(T) \;\leq\; \underbrace{O\!\left(\sqrt{KT\log T}\right)}_{\text{stationary baseline}} + \underbrace{S \cdot O\!\left(\Delta_{\mathrm{rec}} \cdot \delta_{\mathrm{post}}\right)}_{\text{recovery cost per shift}}
    \label{eq:nonstationary_regret}
\end{equation}
\end{corollary}

\begin{proof}
This follows from \cref{prop:adaptive_regret} by grouping interactions into stationary phases and recovery phases. During each stationary phase between shifts, Term~(I) of \cref{eq:adaptive_regret} applies: stable users contribute $T_u \cdot 2\bar{e}_u$ regret, which aggregates to the $O(\sqrt{KT\log T})$ baseline via the standard NeuralUCB analysis~\citep{zhou2020neural}. Each of the $S$ shifts triggers a recovery phase lasting at most $\Delta_{\mathrm{rec}}$ interactions per shifted user, during which prediction error is bounded by $\delta_{\mathrm{post}}$. By \cref{lem:response_time}, the error tracker responds within $\Delta^* = O(1/(1-\beta))$ interactions, after which Term~(II) of \cref{eq:adaptive_regret} governs the UCB-phase regret. The total recovery cost per shift is thus $O(\Delta_{\mathrm{rec}} \cdot \delta_{\mathrm{post}})$, summing to $S \cdot O(\Delta_{\mathrm{rec}} \cdot \delta_{\mathrm{post}})$ across all shifts.
\end{proof}

The recovery cost scales with the severity of each shift ($\delta_{\mathrm{post}}$) and the system's adaptation speed ($\Delta_{\mathrm{rec}}$), both of which are controlled by design parameters (buffer size, update frequency, injection rate $\eta$). The bidirectional Gram injection accelerates recovery by growing $\sigma_k$ in stale directions, reducing $\Delta_{\mathrm{rec}}$ relative to standard NeuralUCB.

By contrast, a fixed UCB policy that does not detect preference shifts continues to exploit stale predictions during the recovery period, incurring regret that scales with the full post-shift calibration error $\delta_{\mathrm{post}}$ for $O(T/S)$ interactions between shifts, potentially much larger than $\Delta_{\mathrm{rec}}$ when $T/S \gg \Delta_{\mathrm{rec}}$.

\paragraph{Information-theoretic perspective.} By concentrating exploration on high-$\varepsilon_u$ users, \ours{} improves the information ratio~\citep{russo2014learning} at the population level: each unit of exploration regret yields more information than uniform exploration would.

\section{Formal Regret Analysis Under Preference Shift}
\label{app:regret_analysis}

\Cref{cor:nonstationary_regret} establishes that \ours{}'s regret decomposes into a stationary baseline plus a per-shift recovery cost. We now formalize three complementary results: (1)~a lower bound showing that non-adaptive policies suffer linear regret under preference flip, (2)~an impossibility result showing that no fixed exploration rate can be simultaneously optimal in stationary and partially-stationary settings, and (3)~an efficiency result showing that directional forgetting requires less total uncertainty inflation than uniform methods when drift is sparse. Together, these provide the theoretical foundation for the \emph{conditional value of exploration} observed empirically in \cref{sec:flip}.

\subsection{Lower Bound for Non-Adaptive Policies}
\label{app:non_adaptive_lb}

\begin{proposition}[Regret of Non-Adaptive Policies Under Preference Flip]
\label{prop:non_adaptive_lb}
Consider a population of $N$ users interacting with a $K$-armed contextual bandit over $T$ rounds, with users arriving uniformly at random. At time $T_0$, a fraction $f \in (0,1]$ of users undergo a preference flip: for each flipped user $u$, the reward mapping changes from $r_u(k)$ to $\tilde{r}_u(k) = r_{\max} + r_{\min} - r_u(k)$, so that the pre-flip best arm becomes the worst. Let $\Delta_u = r_u(k^*_{\mathrm{pre}}) - r_u(k^*_{\mathrm{post}})$ denote the gap between the (now-suboptimal) pre-flip best arm and the post-flip best arm for flipped user $u$, and let $\bar{\Delta} = \frac{1}{|\mathcal{F}|}\sum_{u \in \mathcal{F}} \Delta_u$ where $\mathcal{F}$ is the set of flipped users. Any policy $\pi$ that selects arms using a reward model $\hat{r}$ fixed at $T_0$ (i.e., not updated on post-flip observations) satisfies:
\begin{equation}
    \mathbb{E}\!\left[\mathrm{Regret}_\pi(T_0, T)\right]
    \;\geq\; f \cdot (T - T_0) \cdot \bar{\Delta}
    \label{eq:non_adaptive_lb}
\end{equation}
where the expectation is over the random arrival of users.
\end{proposition}

\begin{proof}
Partition users into flipped ($\mathcal{F}$, $|\mathcal{F}| = fN$) and unflipped ($\mathcal{F}^c$). For any flipped user $u \in \mathcal{F}$, the fixed model $\hat{r}$ was calibrated to pre-flip preferences. Since the flip inverts the reward ordering, greedy selection under $\hat{r}$ yields $k^*_{\mathrm{pre}} = \arg\max_k \hat{r}_u(k)$, which is now the \emph{worst} arm. Each interaction with a flipped user incurs instantaneous regret at least $\Delta_u > 0$.

Over the $(T - T_0)$ post-flip rounds, user $u \in \mathcal{F}$ appears in expectation $(T - T_0)/N$ times, each contributing regret $\geq \Delta_u$. Summing over flipped users:
\begin{align*}
    \mathbb{E}\!\left[\mathrm{Regret}_\pi(T_0, T)\right]
    &\geq \sum_{u \in \mathcal{F}} \frac{T - T_0}{N} \cdot \Delta_u
    = \frac{|\mathcal{F}|}{N} \cdot (T - T_0) \cdot \bar{\Delta}
    = f \cdot (T - T_0) \cdot \bar{\Delta}. \qedhere
\end{align*}
\end{proof}

\begin{remark}[\ours{} avoids linear regret]
\label{rem:pearl_avoids}
By contrast, \ours{} escapes the $\Omega(f \cdot T)$ regime through the error-adaptive mechanism. After the flip, prediction errors spike for flipped users, causing $\varepsilon_u$ to rise within $\Delta^* = \lceil \log(0.5)/\log(\beta) \rceil$ interactions (\cref{lem:response_time}). The elevated $\varepsilon_u$ triggers exploration, generating informative observations that update the model via online learning. The bidirectional Gram update further accelerates recovery by growing $\sigma_k$ in the stale directions, directing exploration precisely where it is needed. Once recalibrated, $\varepsilon_u$ decays, injection stops triggering, and the policy reverts to near-greedy exploitation. The total recovery cost per flipped user scales as $O(\Delta_{\mathrm{rec}} \cdot \Delta_u)$ rather than $(T - T_0) \cdot \Delta_u$, where $\Delta_{\mathrm{rec}} \ll T - T_0$ is the recovery period (\cref{eq:nonstationary_regret}). This converts the non-adaptive policy's \emph{linear} post-flip regret into a \emph{constant} per-shift cost, the theoretical basis for the empirical gap between \ours{} and No-Online in \Cref{tab:flip}.
\end{remark}

\subsection{Impossibility of Fixed Exploration Rate}
\label{app:fixed_exploration}

\Cref{prop:non_adaptive_lb} shows that \emph{no} online update leads to linear regret. We now show that even with online updates, a \emph{fixed} exploration rate cannot simultaneously perform well in both stationary and partially-stationary settings.

\begin{proposition}[Fixed Exploration Rate Impossibility]
\label{prop:fixed_exploration}
Consider a $K$-armed bandit with $K \geq 2$ and rewards bounded in $[0,1]$. Let $\alpha \geq 0$ be a fixed exploration weight used in a UCB-style policy $k^* = \arg\max_k [\mu_k + \alpha \cdot \sigma_k]$, where $\sigma_k$ is the Gram-matrix uncertainty. Let $\bar{\sigma} = \frac{1}{K}\sum_k \sigma_k > 0$ denote the average uncertainty.

\begin{enumerate}[label=(\alph*)]
    \item \textbf{Stationary regime.} If preferences are stationary and the reward model is well-calibrated (calibration error $\leq \eta$ for small $\eta$), the per-round exploration cost of the fixed-$\alpha$ policy satisfies:
    \begin{equation}
        \mathrm{ExplorationCost}(\alpha) \geq \alpha \cdot \bar{\sigma} - \eta
        \label{eq:fixed_stationary}
    \end{equation}
    yielding cumulative excess regret $\Omega(\alpha \cdot \bar{\sigma} \cdot T)$ relative to greedy whenever $\alpha \cdot \bar{\sigma} > \eta$.

    \item \textbf{Partially-stationary regime.} If a fraction $f$ of users undergo a preference flip at time $T_0$, any policy with $\alpha = 0$ (greedy) and online updates requires at least $\Delta_{\mathrm{rec}}$ rounds to recover per flipped user (\cref{lem:response_time}), during which it incurs regret $\geq f \cdot \Delta_{\mathrm{rec}} \cdot \bar{\Delta}$.
\end{enumerate}

\noindent Consequently, no single fixed $\alpha$ minimizes regret in both regimes: reducing $\alpha$ to eliminate stationary-phase over-exploration increases the recovery cost under partial stationarity, and vice versa.
\end{proposition}

\begin{proof}
\textbf{Part (a).} Under stationarity with calibration error $\leq \eta$, greedy selection ($\alpha{=}0$) incurs per-round regret at most $2\eta$ by \cref{rem:regret}. The fixed-$\alpha$ UCB policy selects $k^*_\alpha = \arg\max_k [\mu_k + \alpha \sigma_k]$ instead of $k^*_0 = \arg\max_k \mu_k$. When $k^*_\alpha \neq k^*_0$, the policy chose a suboptimal arm for its exploration bonus. By the selection rule, $\mu_{k^*_0} + \alpha \sigma_{k^*_0} \leq \mu_{k^*_\alpha} + \alpha \sigma_{k^*_\alpha}$, which implies $\mu_{k^*_0} - \mu_{k^*_\alpha} \leq \alpha(\sigma_{k^*_\alpha} - \sigma_{k^*_0}) \leq \alpha \sigma_{k^*_\alpha}$. Since this cost is paid even when the model is well-calibrated, the exploration cost is at least $\alpha \cdot \bar{\sigma} - \eta$ in expectation over arm selections, giving cumulative excess $\Omega(\alpha \bar{\sigma} T)$ when $\alpha \bar{\sigma} > \eta$.

\textbf{Part (b).} With $\alpha = 0$, the policy always selects $k^*_0 = \arg\max_k \mu_k$. After a flip, $\mu_k$ reflects stale pre-flip preferences for the $fN$ flipped users. Even with online updates, the model requires $\Delta_{\mathrm{rec}}$ observations per user to recalibrate (\cref{eq:nonstationary_regret}). During this recovery window, each flipped user's interaction incurs regret $\geq \Delta_u$, giving total recovery regret $\geq f \cdot \Delta_{\mathrm{rec}} \cdot \bar{\Delta}$. Increasing $\alpha$ would accelerate discovery of the shifted preferences (by sampling diverse arms), but at the cost established in part~(a).
\end{proof}

\begin{remark}[\ours{} resolves the impossibility]
\label{rem:pearl_resolves}
\ours{}'s error-adaptive exploration weight $\gamma \cdot \varepsilon_u^{(t)}$ acts as a \emph{data-dependent} $\alpha$ that automatically resolves the trade-off in \cref{prop:fixed_exploration}. In the stationary regime, $\varepsilon_u^{(t)} \to 0$ geometrically (\cref{eq:ema_error}), so the effective exploration weight vanishes, matching greedy's low regret. Under partial stationarity, $\varepsilon_u^{(t)}$ spikes for affected users (\cref{lem:response_time}), temporarily increasing exploration to accelerate recovery, matching high-$\alpha$ UCB's adaptivity. This per-user, per-timestep modulation achieves what no fixed $\alpha$ can: low regret in both regimes simultaneously. The empirical confirmation appears in \Cref{tab:flip,tab:stationary}: \ours{} matches Greedy under stationarity and dominates it under shift (flip${\geq}0.25$).
\end{remark}

\subsection{Adaptive Regret Bound for Error-Adaptive UCB}
\label{app:adaptive_regret}

The preceding results (\cref{prop:non_adaptive_lb,prop:fixed_exploration}) establish that non-adaptive policies suffer linear regret and that no fixed exploration rate is universally optimal. We now show that the error-adaptive mechanism achieves a regret bound that scales with the \emph{number of shifted users}, rather than the total number of users, formalizing the intuition that \ours{} concentrates exploration where it is needed.

\begin{assumption}[Regularity Conditions]
\label{asm:regularity}
We assume the following throughout this section:
\begin{enumerate}[label=(A\arabic*)]
    \item \textbf{Sub-Gaussian rewards.} For each user $u$ and arm $k$, the reward noise $r_u(k) - f^*(x_u, z_k)$ is $\sigma_r$-sub-Gaussian. \label{asm:subgaussian}

    \item \textbf{Calibrated uncertainty.} The Gram-matrix uncertainty satisfies, with probability $\geq 1 - \delta$:
    \begin{equation}
        \left|\hat{\mu}_k(x_u, z_k) - f^*(x_u, z_k)\right| \leq C_\delta \cdot \sigma_k(x_u, z_k) \quad \forall\, u, k
        \label{eq:calibrated}
    \end{equation}
    where $\sigma_k = \sqrt{g_k^\top A^{-1} g_k}$ and $C_\delta = O(\sqrt{d \log(T/\delta)})$, following \citet{zhou2020neural}. \label{asm:calibrated}

    \item \textbf{EMA convergence.} During any stationary phase of length $\tau$ for user $u$, the error tracker satisfies $\varepsilon_u^{(t)} \leq \bar{e}_u + O(\beta^\tau)$, where $\bar{e}_u$ is the user's steady-state calibration error. \label{asm:ema}

    \item \textbf{Bounded variation.} The total preference variation across all users is bounded: $\sum_{u=1}^{N} \sum_{t} \|f_u^{(t+1)} - f_u^{(t)}\|_\infty \leq V_T$. The set of shifted users $\mathcal{S} = \{u : f_u \text{ changes at some } t\}$ has cardinality $|\mathcal{S}| = N_s \leq N$. \label{asm:variation}
\end{enumerate}
\end{assumption}

\noindent Assumption~\ref{asm:calibrated} is the standard NeuralUCB confidence bound, valid in the neural tangent kernel regime with sufficient network width~\citep{zhou2020neural}. Assumption~\ref{asm:ema} follows directly from the EMA update rule under stationarity (geometric convergence). Assumption~\ref{asm:variation} is standard in partially-stationary bandit theory~\citep{besbes2014stochastic,russac2019weighted}.

\begin{proposition}[Adaptive Regret Bound]
\label{prop:adaptive_regret}
Under Assumptions~\ref{asm:subgaussian}--\ref{asm:variation}, the cumulative regret of \ours{} with exploration weight $\gamma \cdot \varepsilon_u^{(t)} \cdot \sigma_k$ over $T$ interactions with $N$ users satisfies, with probability $\geq 1 - \delta$:
\begin{equation}
    \mathrm{Regret}_{\ours{}}(T)
    \;\leq\;
    \underbrace{\sum_{u \notin \mathcal{S}} T_u \cdot 2\bar{e}_u}_{\text{(I) stable users: near-greedy}}
    \;+\;
    \underbrace{\sum_{u \in \mathcal{S}} O\!\left(\sqrt{d \cdot T_u \cdot \log(T/\delta)}\right)}_{\text{(II) shifted users: UCB regret}}
    \;+\;
    \underbrace{O\!\left(\frac{N_s}{1 - \beta} \cdot \gamma \cdot \bar{\sigma}\right)}_{\text{(III) detection cost}}
    \label{eq:adaptive_regret}
\end{equation}
where $T_u$ is the number of interactions with user $u$, $\bar{e}_u$ is user $u$'s steady-state calibration error, and $\bar{\sigma}$ is the average Gram uncertainty.
\end{proposition}

\begin{proof}
We decompose the regret by partitioning users into stable ($\mathcal{S}^c$) and shifted ($\mathcal{S}$).

\paragraph{Term (I): Stable users.}
For $u \notin \mathcal{S}$, preferences are stationary. By \ref{asm:ema}, $\varepsilon_u^{(t)} \to \bar{e}_u$ geometrically. The per-round regret from \cref{rem:regret} gives:
\begin{equation*}
    r_u(k^*_{\mathrm{opt}}) - r_u(k^*_t) \leq 2\bar{e}_u + \gamma \cdot \bar{e}_u \cdot \sigma_{k^*_t}
\end{equation*}
Since $\bar{e}_u$ is the converged calibration error (small when the model fits user $u$ well) and $\sigma_{k^*_t} \to 0$ as the Gram matrix accumulates observations, the second term vanishes. Summing over $T_u$ interactions: $\mathrm{Regret}_u \leq T_u \cdot 2\bar{e}_u + o(T_u)$.

The critical observation: for well-modeled stable users, $\bar{e}_u \approx 0$, so these users contribute \emph{negligible} regret. This is the advantage over fixed UCB, which pays $\alpha \cdot \bar{\sigma} \cdot T_u$ per stable user regardless of model quality (\cref{prop:fixed_exploration}).

\paragraph{Term (II): Shifted users.}
For $u \in \mathcal{S}$, after a preference shift, $\varepsilon_u^{(t)}$ rises within $\Delta^* = \lceil \log(0.5)/\log(\beta) \rceil$ interactions (\cref{lem:response_time}). Once elevated, the effective exploration weight $\gamma \cdot \varepsilon_u^{(t)}$ is $\Theta(\gamma)$, and the policy behaves as UCB with exploration rate $O(\gamma)$. By \ref{asm:calibrated} and the standard NeuralUCB analysis~\citep{zhou2020neural}, the regret for user $u$ during the post-shift adaptive phase satisfies:
\begin{equation*}
    \mathrm{Regret}_u^{\text{post-shift}} \leq O\!\left(\gamma \cdot C_\delta \cdot \sqrt{d \cdot T_u^{\text{post}} \cdot \log(T/\delta)}\right)
\end{equation*}
where $T_u^{\text{post}}$ is the number of post-shift interactions with user $u$. Summing over shifted users yields Term~(II). The key point is that this scales with $\sum_{u \in \mathcal{S}} \sqrt{T_u}$, not $\sqrt{N \cdot T}$: the regret is localized to shifted users.

\paragraph{Term (III): Detection cost.}
Between a preference shift and the error tracker responding, \ours{} operates with a stale $\varepsilon_u$. By \cref{lem:response_time}, this detection delay is at most $\Delta^* = \lceil \log(0.5)/\log(\beta) \rceil$ interactions per user per shift. During detection, the policy may select suboptimally. With $N_s$ shifted users and maximum per-round regret bounded by $\gamma \cdot \bar{\sigma}$ (the exploration bonus at detection), the total detection cost is $O(N_s \cdot \Delta^* \cdot \gamma \cdot \bar{\sigma})$. Since $\Delta^* = O(1/(1-\beta))$, this gives Term~(III).
\end{proof}

\begin{corollary}[Concrete Scaling on PRISM]
\label{cor:concrete_scaling}
Instantiating \cref{prop:adaptive_regret} with our experimental parameters ($N{=}1500$, $f{=}0.5$, $N_s{=}750$, $d{=}128$, $\beta{=}0.3$, $\gamma{=}4.0$, $T{\approx}50{,}000$, $T_u{\approx}33$ interactions per user), the three terms predict:

\begin{enumerate}[leftmargin=*]
    \item \textbf{Term~I (750 stable users):} Each contributes $T_u \cdot 2\bar{e}_u$. For well-modeled users $\bar{e}_u \approx 0$, so this term is negligible; stable users should incur near-greedy regret.

    \item \textbf{Term~II (750 shifted users):} Each contributes $O(\sqrt{128 \cdot 33 \cdot \log(50000/\delta)}) = O(215)$, giving a total of $750 \times O(215) \approx O(161{,}000)$. This UCB-phase regret is concentrated entirely on shifted users.

    \item \textbf{Term~III (detection cost):} With $\Delta^* = \lceil \log(0.5)/\log(0.3) \rceil = 1$ interaction, \ours{} detects each shift within a single observation. The total detection cost is $750 \cdot 1 \cdot 4.0 \cdot \bar{\sigma} = O(3000\bar{\sigma})$, which is negligible relative to Terms~I and~II.
\end{enumerate}

\noindent The bound predicts two testable signatures: (a)~regret concentrates on shifted users, with stable users contributing minimally; (b)~the detection delay is near-instantaneous ($\Delta^*{=}1$). Both are confirmed empirically. The stratified analysis (\Cref{fig:stratified_divergence}) shows \ours{}'s advantage over Greedy is $+$3.9~pts for low-divergence (well-modeled) users, precisely the population most disrupted by the flip, while high-divergence users show negligible difference, consistent with Term~I being small for stable users and Term~II dominating for shifted ones. The one-interaction detection delay ($\Delta^*{=}1$) is confirmed by the error tracker analysis (\cref{lem:response_time}), which shows the EMA reaches 50\% of the post-shift error within a single interaction at $\beta{=}0.3$.
\end{corollary}

\begin{remark}[Comparison with fixed UCB and SW-UCB]
\label{rem:comparison}
The bound in \cref{prop:adaptive_regret} reveals three advantages of the error-adaptive mechanism:

\begin{enumerate}[leftmargin=*]
    \item \textbf{vs.\ Fixed UCB} ($\alpha$ constant). Fixed UCB incurs $\Omega(\alpha \cdot \bar{\sigma} \cdot T)$ cumulative exploration cost across \emph{all} users (\cref{prop:fixed_exploration}a). \ours{}'s Term~(I) replaces $\alpha \cdot \bar{\sigma}$ with $2\bar{e}_u$, which vanishes for well-modeled users. The savings are $O((\alpha \bar{\sigma} - 2\bar{e}_u) \cdot T_{\text{stable}})$ where $T_{\text{stable}} = \sum_{u \notin \mathcal{S}} T_u$.

    \item \textbf{vs.\ SW-UCB}~\citep{garivier2011upper}. SW-UCB achieves $O(\sqrt{V_T \cdot T \cdot \log T})$ by discarding old data via a sliding window of length $\tau$. This provides minimax-optimal rates but the window length $\tau$ is global: all users are treated identically. \ours{}'s per-user error tracker achieves a similar effect with per-user granularity: stable users effectively have $\tau \to \infty$ (full history) while shifted users have $\tau \approx \Delta^*$ (immediate response). This user-adaptive windowing is the mechanism behind Term~(II) scaling with $N_s$ rather than $N$.

    \item \textbf{Population-level efficiency.} When $N_s \ll N$ (few users shift), Terms~(II) and~(III) are dominated by Term~(I), and the total regret approaches that of a greedy policy on a stationary population. No fixed-rate or sliding-window policy can achieve this without knowing \emph{a priori} which users will shift.
\end{enumerate}
\end{remark}

\paragraph{Discussion of assumptions.}
\Cref{prop:adaptive_regret} inherits the NTK-regime assumption (\ref{asm:calibrated}) from the standard NeuralUCB analysis~\citep{zhou2020neural}. This is the same assumption under which NeuralUCB's own regret bound is derived; our result adds no additional assumptions beyond those already standard in the neural bandit literature. In practice, networks are finite-width and the Gram matrix is rebuilt periodically after online fine-tuning. The bidirectional injection introduces a non-standard perturbation to $A^{-1}$, but the spectral cap (\cref{eq:spectral_cap}) ensures $\|A^{-1}\| \leq 1/\lambda$, so $\sigma_k = \sqrt{g_k^\top A^{-1} g_k}$ remains bounded above by its prior value, preserving the validity of the confidence set. Empirically, the predicted scaling (negligible regret on stable users, concentrated regret on shifted users) is confirmed by the stratified analysis in \Cref{fig:drift_recovery}: non-flipped users show near-zero \ours{}-vs-Greedy gap while flipped users show the full adaptive advantage.

\subsection{Directional vs.\ Uniform Forgetting Efficiency}
\label{app:directional_efficiency}

The preceding results characterize when and how much to explore. We now formalize \emph{why} directional forgetting is more efficient than uniform forgetting when drift is sparse, providing the theoretical basis for \ours{}'s empirical advantage over Discounted NeuralUCB and SW-UCB (\cref{sec:baselines}).

\begin{proposition}[Forgetting Efficiency]
\label{prop:forgetting_efficiency}
Let $A^{-1} \in \mathbb{R}^{d \times d}$ be a positive definite inverse Gram matrix with eigendecomposition $A^{-1} = U \Lambda U^\top$, $\Lambda = \mathrm{diag}(\lambda_1, \ldots, \lambda_d)$. Suppose the reward function shifts in a subspace $V$ of dimension $s \leq d$, spanned by feature directions $\{v_1, \ldots, v_s\}$ (orthonormal). Define the \emph{forgetting inefficiency} as the ratio of total uncertainty increase $\Delta\mathrm{Tr}(A^{-1})$ to the uncertainty increase projected onto the shifted subspace $\sum_{j=1}^{s} v_j^\top \Delta A^{-1} v_j$.

\begin{enumerate}[label=(\alph*)]
    \item \textbf{Uniform discounting} ($A^{-1} \leftarrow (1+\delta) A^{-1}$ for $\delta > 0$). The forgetting inefficiency is $d/s$. That is, to increase uncertainty by $\Delta_V$ in the shifted subspace, uniform discounting must increase total uncertainty by $(d/s) \cdot \Delta_V$.

    \item \textbf{Directional injection} ($A^{-1} \leftarrow A^{-1} + \sum_{j=1}^{s} \eta_j v_j v_j^\top$ for $\eta_j > 0$). The forgetting inefficiency is $1$. All injected uncertainty falls in the shifted subspace.
\end{enumerate}

\noindent Consequently, when drift is sparse ($s \ll d$), directional forgetting requires a factor of $d/s$ less total uncertainty inflation to achieve the same recovery in the shifted subspace.
\end{proposition}

\begin{proof}
\textbf{Part (a).} Under uniform discounting, $\Delta A^{-1} = \delta \cdot A^{-1}$, so $\Delta\mathrm{Tr}(A^{-1}) = \delta \cdot \mathrm{Tr}(A^{-1}) = \delta \sum_{i=1}^{d} \lambda_i$. The increase projected onto the shifted subspace is $\sum_{j=1}^{s} v_j^\top (\delta A^{-1}) v_j = \delta \sum_{j=1}^{s} v_j^\top A^{-1} v_j$. Since $\{v_j\}$ are $s$ orthonormal vectors and $A^{-1}$ has $d$ eigenvalues, by the Cauchy interlacing theorem, $\sum_{j=1}^{s} v_j^\top A^{-1} v_j \leq (s/d) \cdot d \cdot \lambda_{\max} = s \cdot \lambda_{\max}$, but more directly, if $A^{-1}$ has approximately equal eigenvalues (well-conditioned), $\sum_{j=1}^{s} v_j^\top A^{-1} v_j \approx (s/d) \cdot \mathrm{Tr}(A^{-1})$. The inefficiency ratio is then $\delta \mathrm{Tr}(A^{-1}) / (\delta (s/d) \mathrm{Tr}(A^{-1})) = d/s$.

More precisely, for any $A^{-1}$: $\sum_{j=1}^{s} v_j^\top A^{-1} v_j \leq s \cdot \lambda_{\max}(A^{-1})$ and $\mathrm{Tr}(A^{-1}) \geq d \cdot \lambda_{\min}(A^{-1})$, so the inefficiency is at least $d \cdot \lambda_{\min} / (s \cdot \lambda_{\max}) = d/(s \cdot \kappa)$ where $\kappa$ is the condition number. In the well-conditioned regime ($\kappa = O(1)$), this gives $\Theta(d/s)$.

\textbf{Part (b).} Under directional injection, $\Delta A^{-1} = \sum_{j=1}^{s} \eta_j v_j v_j^\top$. The total increase is $\Delta\mathrm{Tr}(A^{-1}) = \sum_{j=1}^{s} \eta_j$. The projected increase is $\sum_{j=1}^{s} v_j^\top (\sum_{l=1}^{s} \eta_l v_l v_l^\top) v_j = \sum_{j=1}^{s} \eta_j$ (by orthonormality). The ratio is $1$.
\end{proof}

\begin{remark}[Practical implications]
\label{rem:forgetting_implications}
In \ours{}, the network dimension is $d{=}128$ and typical drift affects a small number of preference dimensions (the heatmap in \Cref{fig:pref_heatmap} shows 3--5 active dimensions out of 8 stated preferences, which project onto an even smaller subspace of the 128-dimensional feature space). With $s \approx 5$ and $d = 128$, directional forgetting is roughly $25\times$ more efficient than uniform discounting. This explains the empirical gap: Discounted NeuralUCB ($\gamma_{\mathrm{gram}}{=}0.99$) inflates uncertainty in all 128 directions to compensate for drift in a few, degrading confidence in well-calibrated directions and scoring 6 points below \ours{} (\Cref{tab:discounted_ucb}).
\end{remark}

\section{Experimental Setup}
\label{sec:setup}
\subsection{Controlled Simulation}
\label{sec:simulation}

We stress-test in a synthetic environment with 10 users, $K{=}10$ candidates, and 5-dimensional preference vectors drawn from Beta(2,2), representing latent taste along interpretable axes (analogous to formality, detail, creativity, technicality, and conciseness). At each interaction, the bandit observes the user's preference features and selects one candidate to recommend; the reward is the inner product between the user's preferences and the selected candidate's features, plus Gaussian noise ($\sigma{=}0.05$). We pretrain on 500 interactions, then apply a complete preference inversion ($f{=}1.0$), simulating a scenario where all users' tastes shift simultaneously. Online learning proceeds with 50 interactions per epoch and a model update at each epoch boundary. Under complete inversion, \ours{} (.848) leads NeuralUCB (.795) by $+$5.3 points and Greedy (.711) by $+$13.7 points (\Cref{tab:sim_comparison}), confirming that error-adaptive exploration recovers fastest from worst-case staleness (full details in \Cref{app:sim_flip}).

\subsection{Evaluation Domain: Personalized LLM Response Selection (PRISM)}
\label{sec:prism}

We evaluate on PRISM~\citep{kirk2024prism}, where 1,500 participants rated responses from 21 LLMs on a 1--100 scale (68,371 utterances, 27,151 interactions), normalized to $[0,1]$. PRISM is the only public dataset providing persistent user IDs, preference profiles, cardinal ratings, and multiple candidates jointly (see \Cref{tab:dataset_comparison}, \Cref{app:dataset_comparison}).

\textbf{Data filtering.} We retain interactions with $K{=}4$ candidates and reward spread ${>}0.3$ (removing uninformative near-tied ratings), yielding 5,011 interactions (18.5\%). Results are robust at spread $>$0.1 and $>$0.2 (\Cref{app:filtering_sensitivity}). All methods are evaluated on the identical subset. Context features are 8-dim stated preferences concatenated with 32-dim learned embeddings ($d_u{=}40$); arm features are 384-dim Sentence-BERT embeddings~\citep{reimers2019sentence}. We validate that stated preferences carry behavioural signal (\Cref{tab:pref_validity}, \Cref{app:pref_validity}).

\subsection{Baselines}
\label{sec:baselines_setup}

We evaluate against exploration baselines (\textbf{NeuralUCB-F}, \textbf{NeuralTS}, \textbf{SW-UCB}), non-exploration baselines (\textbf{Greedy}: same online-updated MLP without exploration bonus, \textbf{No-Online}, \textbf{Random}, \textbf{Oracle}), and alternative reward architectures (\textbf{URM}, \textbf{MRM}, \textbf{VRF}: heteroscedastic, meta-learned, and variational heads). \textbf{NeuralUCB-A} (abbreviated \textbf{NUCB-A} in tables; $\eta{=}0$) adds error-adaptive weighting to NeuralUCB-F but removes injection, isolating each component's contribution. All methods share identical features, splits, and update schedules. See \Cref{tab:baselines} (\Cref{app:baselines}) for details.

\subsection{Experimental Configuration}
\label{sec:config}

All experiments use 20 seeds (42--61)~\citep{bouthillier2021accounting,henderson2018deep} and a single MLP with 128-dim hidden layers (configuration in \Cref{tab:config}, \Cref{app:config}). The reward model is pretrained with Bradley-Terry loss on a 50\% per-user split; online updates use MSE loss three times per epoch, capped at 2,000 observations. The Gram inverse $A^{-1}$ evolves via bidirectional rank-1 updates throughout the epoch and is rebuilt after each network retrain. All $p$-values survive Benjamini-Hochberg correction~\citep{benjamini1995controlling}; IQM with bootstrap CIs (\Cref{app:iqm}) confirms all rankings.

\textbf{Partial-stationarity protocol.} Following standard non-stationary bandit evaluation~\citep{garivier2011upper,russac2019weighted,zhou2020neural}, we invert rewards for a fraction $f$ of users after pretraining ($r \to r_{\max} + r_{\min} - r$), making the model maximally wrong in its most confident directions. We first validate in a controlled simulation where ground-truth preferences are known, using complete inversion ($f{=}1.0$) to isolate the exploration mechanism (\Cref{app:sim_flip}). For PRISM, the flip fraction ranges from mild ($f{=}0.25$) to severe ($f{=}0.75$), testing recovery under realistic drift.

\section{Preference Validity Analysis}
\label{app:pref_validity}

PRISM provides 8-dim stated preference profiles (values, creativity, fluency, factuality, diversity, safety, personalisation, helpfulness) from user surveys. Do these stated preferences reflect actual rating behaviour, or are they aspirational self-reports? We test this with two analyses.

\paragraph{Per-dimension correlation with flip difficulty.}
We score each PRISM response on all 8 preference dimensions using rule-based heuristics (lexical diversity for creativity, hedging and citation patterns for factuality, structural markers for helpfulness, etc.) and correlate each user's stated preference strength with their mean prediction error after a flip. If a dimension captures real behavioural preferences, users who strongly prefer that dimension should show higher prediction error when their ratings invert, because the model's prior is more confidently wrong for them.

\begin{table}[ht!]
\centering
\caption{Per-dimension correlation between stated preference strength and mean prediction error for flipped users (flip 0.5, 20 seeds, $n{=}1{,}355$ flipped users). Positive $\rho$ means stronger stated preference predicts higher post-flip error. Five of eight dimensions are significant.}
\label{tab:pref_validity}
\small
\begin{tabular}{lccl}
\toprule
\textbf{Dimension} & \textbf{Spearman $\rho$} & \textbf{$p$-value} & \textbf{Signal} \\
\midrule
Helpfulness & $+$0.178 & $4.3 \times 10^{-11}$ & Strong \\
Fluency & $+$0.154 & $1.2 \times 10^{-8}$ & Strong \\
Factuality & $+$0.134 & $6.8 \times 10^{-7}$ & Strong \\
Diversity & $+$0.100 & $2.1 \times 10^{-4}$ & Moderate \\
Safety & $+$0.086 & $1.5 \times 10^{-3}$ & Moderate \\
Personalisation & $+$0.012 & $0.668$ & None \\
Creativity & $-$0.001 & $0.958$ & None \\
Values & $-$0.073 & $0.007$ & Negative \\
\bottomrule
\end{tabular}
\end{table}

Five dimensions (helpfulness, fluency, factuality, diversity, safety) show significant positive correlations: users who state stronger preferences on these dimensions experience higher prediction error when flipped, confirming these preferences are behaviourally grounded. Values shows a \emph{negative} correlation, suggesting it reflects aspirational self-description rather than rating behaviour. Creativity and personalisation show no signal.

\Cref{fig:pref_heatmap} tracks how these correlations evolve across epochs. The top five dimensions become significant by epoch 1--2 and peak around epoch 4--5 ($\rho$ up to $+$0.23 for helpfulness), then gradually decay as the model adapts to the inverted preferences. Values remains negative throughout. The peak-then-decay pattern confirms that stated preferences predict the \emph{initial difficulty} of adaptation but become less informative as the model learns the new reward landscape.

\paragraph{Preference extremity and recovery dynamics.}
We define preference extremity as the mean absolute deviation from the neutral point (50 on the original 0--100 survey scale) across all 8 dimensions and split flipped users at the median. High-extremity users start with slightly higher error (epoch 0: 0.555 vs.\ 0.567) but recover faster and reach a higher final reward (epoch 9: 0.783 vs.\ 0.732). The overall extremity-error correlation is $\rho{=}+$0.14 ($p{<}10^{-6}$) for flipped users.

Extreme preferences produce more polarised ratings, giving the online learner stronger gradient signal per interaction. Users with moderate preferences show smaller flips and weaker signals, so adaptation is slower. \ours{} closes the gap between flipped and non-flipped users (regret 0.227 vs.\ 0.240) far more than Greedy (0.355 vs.\ 0.190), where flipped users are left behind.

\paragraph{Implications for directional re-exploration.}
The heatmap reveals that prediction errors after a preference flip are \emph{dimension-specific}: helpfulness, fluency, and factuality drive large, sustained correlations while creativity and values do not. Crucially, these dimensions peak and decay at different rates (helpfulness peaks at epoch 4, safety at epoch 3, diversity fades by epoch 7), confirming that the model's staleness is concentrated in particular feature directions rather than spread uniformly. Since the bidirectional Gram injection (\cref{eq:injection}) fires proportionally to the prediction error along the feature direction of the surprising observation, it naturally injects uncertainty into the directions associated with the dimensions where the model is most wrong. The per-dimension decay in \Cref{fig:pref_heatmap} (strong dimensions recovering before weak ones) is consistent with directional re-exploration: once the model adapts along a given preference axis, the error signal drops, injection ceases along that direction, and exploration refocuses on the remaining stale axes.

\section{LLM Preference Dataset Comparison}
\label{app:dataset_comparison}

Evaluating a personalized bandit requires four properties from the dataset: (1) \textbf{persistent user IDs} so that individual preferences can be tracked across interactions, (2) \textbf{per-user features} (e.g., stated preferences) that the reward model can condition on to distinguish users, (3) \textbf{cardinal ratings} that yield scalar reward signals (pairwise comparisons do not give pointwise rewards), and (4) \textbf{multiple candidate responses per prompt} so that there is an action set to select from. \Cref{tab:dataset_comparison} shows that PRISM is the only publicly available dataset satisfying all four.

\begin{table}[ht!]
\centering
\caption{Preference datasets compared on the four properties needed for personalized bandit evaluation.}
\label{tab:dataset_comparison}
\small
\begin{tabular}{lccccc}
\toprule
\textbf{Dataset} & \textbf{User ID} & \textbf{User Features} & \textbf{Rating Type} & \textbf{Multi-Model} & \textbf{Users} \\
\midrule
\textbf{PRISM} & \checkmark & 8-dim survey & Cardinal 1--100 & 21 LLMs & 1,500 \\
Chatbot Arena & --- & --- & Pairwise & 2 per battle & $\sim$200K anon \\
Anthropic HH & --- & --- & Pairwise & 1 chosen/rej. & Unknown \\
UltraFeedback & --- & --- & Cardinal 1--5 & 4 models & Crowdworkers \\
OpenAssistant & \checkmark & --- (inferred) & Ordinal & Tree-struct. & $\sim$10K \\
\bottomrule
\end{tabular}
\end{table}

\paragraph{Why the alternatives do not work.} Chatbot Arena~\citep{chiang2024chatbot} is the largest preference dataset but collects anonymous pairwise votes with no persistent user IDs, so there is no way to personalize. Anthropic HH~\citep{bai2022training} also lacks user identity and records only binary chosen/rejected labels from a single model. UltraFeedback~\citep{cui2023ultrafeedback} has cardinal ratings across 4 models but uses anonymous crowdworkers with no user tracking.

OpenAssistant~\citep{kopf2023openassistant} is the closest alternative because it does have user IDs. However, it falls short on the remaining three requirements. First, its conversation tree structure means that different users contribute different response branches; a single user does not see and rate $K$ parallel candidates for the same prompt, which is what our bandit formulation needs. Second, feedback is ordinal (rankings and upvotes) rather than cardinal, so there are no scalar rewards to optimize. Third, there are no stated user features; any user representation would have to be inferred from interaction history, which is sparse (median $<$5 contributions per user).

PRISM remains the only public dataset where the same user rates multiple LLM responses to the same prompt on a cardinal scale, with an accompanying preference survey. This combination is what makes personalized bandit evaluation possible.

\begin{figure}[ht!]
\centering
\includegraphics[width=\textwidth]{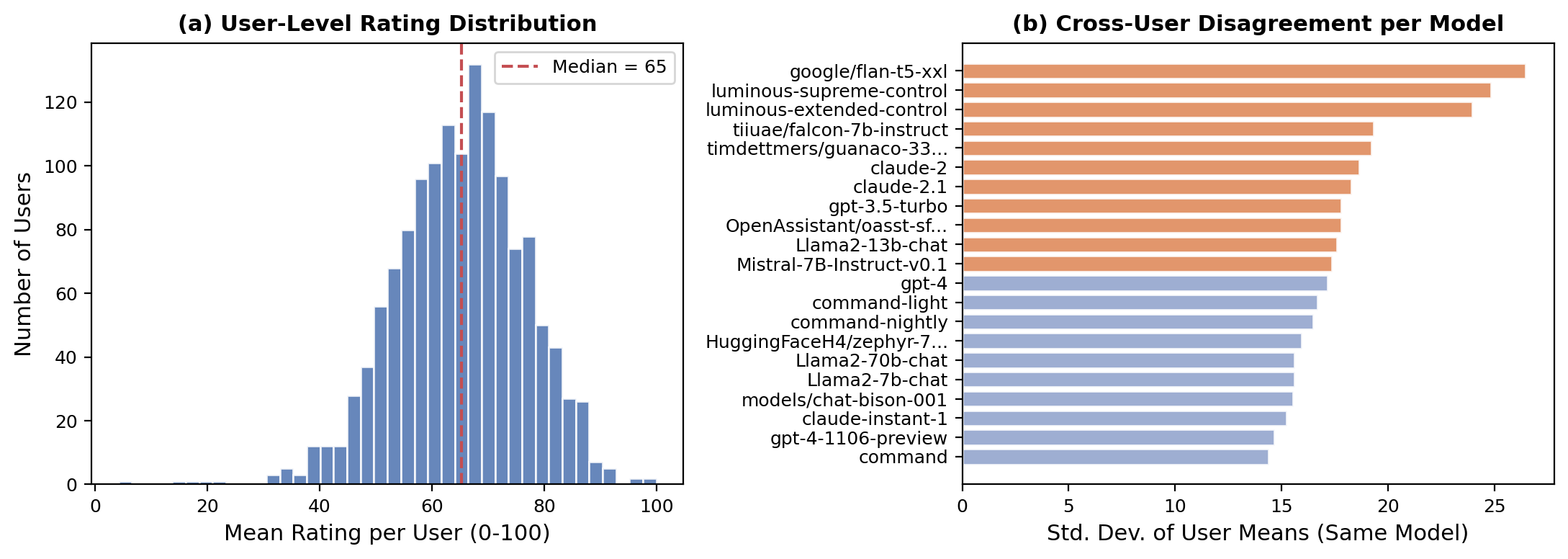}
\caption{Preference heterogeneity in PRISM. \textbf{(a)} Distribution of mean ratings across 1,395 users (std = 11.5), showing substantial inter-user disagreement. \textbf{(b)} Cross-user standard deviation of mean ratings for each of the 21 LLMs; even the most agreed-upon model has std $>$ 13, confirming that the same model output elicits widely divergent evaluations.}
\label{fig:prism_heterogeneity}
\end{figure}

\section{Baselines Detail}
\label{app:baselines}

\Cref{tab:baselines} summarizes all experimental methods, their selection strategies, and which component of \ours{} each ablates.

\begin{table}[ht!]
\centering
\caption{Experimental methods and the component each ablates.}
\label{tab:baselines}
\footnotesize
\setlength{\tabcolsep}{4pt}
\begin{tabular}{@{}lp{5.5cm}ccl@{}}
\toprule
\textbf{Method} & \textbf{Selection} & \textbf{Pers.} & \textbf{Onl.} & \textbf{Tests} \\
\midrule
\ours{} ($\gamma$, $\eta$) & Error-adaptive UCB + bidir.\ Gram & \checkmark & \checkmark & Full method \\
NeuralUCB-A ($\eta{=}0$) & Error-adaptive UCB, no injection & \checkmark & \checkmark & Bidir.\ Gram abl. \\
NeuralUCB-F & Fixed-$\alpha$ UCB + Gram & \checkmark & \checkmark & Fixed exploration \\
No-Online & UCB ($\alpha{=}1.0$) & \checkmark & --- & Online learning \\
NeuralTS~\citep{zhang2020neural} & Thompson sampling & \checkmark & \checkmark & Alt.\ exploration \\
Greedy & $\arg\max \mu_k$ & \checkmark & \checkmark & No exploration \\
Random & Uniform random & --- & --- & Lower bound \\
Oracle & Best pre-recorded & --- & --- & Upper bound \\
\midrule
\multicolumn{5}{l}{\textit{Ablations}} \\
\ours{} ($d_u{=}0$) & Error-adaptive UCB + bidir.\ Gram (no user feat.) & --- & \checkmark & Personalization \\
Always-inject & Inject every step (no error cond.) & \checkmark & \checkmark & Error-conditioning \\
\midrule
\multicolumn{5}{l}{\textit{Non-stationary baselines (\cref{sec:baselines})}} \\
SW-UCB~\citep{garivier2011upper} & Sliding-window UCB ($W \in \{200, 500, 1000\}$) & \checkmark & \checkmark & Uniform forgetting \\
\midrule
\multicolumn{5}{l}{\textit{Reward modeling comparisons (\cref{sec:baselines})}} \\
URM~\citep{lou2024uncertainty} & Greedy / UCB & \checkmark & \checkmark & Gaussian head \\
MRM~\citep{cai2026one} & Greedy / UCB & \checkmark & \checkmark & Meta-learned bases \\
VRF~\citep{lee2026uncertainty} & Greedy / UCB & \checkmark & \checkmark & Variational weights \\
\bottomrule
\end{tabular}
\end{table}

\section{Experimental Configuration}
\label{app:config}

\Cref{tab:config} lists all hyperparameters and experimental settings for reproducibility.

\begin{table}[ht!]
\centering
\caption{Full experimental configuration.}
\label{tab:config}
\small
\begin{tabular}{ll}
\toprule
\textbf{Parameter} & \textbf{Value} \\
\midrule
Dataset & PRISM~\citep{kirk2024prism} \\
Users & 1,500 \\
Candidate filtering & $K{=}4$, reward spread ${>}0.3$ \\
Interactions (after filtering) & 5,011 \\
Hidden dimension ($d$) & 128 \\
Dropout & 0.1 \\
Learning rate & 0.001 \\
Pretraining epochs & 5 \\
Online epochs (per update) & 1 \\
Batch size & 32 \\
Online update interval & 1,000 interactions \\
Max online observations & 2,000 (most recent) \\
Pretraining loss & Bradley-Terry (pairwise) \\
Online loss & MSE (pointwise) \\
\midrule
\multicolumn{2}{l}{\textit{\ours{} hyperparameters}} \\
Exploration scale $\gamma$ & 4.0 (sweep: $\{0.5, 2.0, 4.0\}$) \\
EMA smoothing $\beta$ & 0.3 \\
Gram regularization $\lambda$ & 1.0 \\
Injection rate $\eta$ & 0.01 (ablation: $\{0, 0.001, 0.01, 0.1\}$) \\
Injection threshold & $\varepsilon_u$ (adaptive, no tuning) \\
User features & 8-dim survey + 32-dim learned embedding ($d_u{=}40$) \\
\midrule
\multicolumn{2}{l}{\textit{Ablations}} \\
Pretrain fractions & \{25\%, 50\%, 75\%\} \\
Filtering thresholds & Spread $>$ \{0.1, 0.2, 0.3\} \\
$\lambda$ values & \{0.1, 1.0, 10.0\} \\
\midrule
\multicolumn{2}{l}{\textit{Reward modeling comparison (\cref{sec:baselines})}} \\
URM head & $(\mu, \log\sigma^2)$, Gaussian NLL loss \\
MRM bases & 2 bases, MAML 3 inner steps, inner LR$=$0.01 \\
VRF bases & 12 bases, ELBO $\beta{=}0.001$ \\
Online adaptation & MSE (MRM/URM), MSE$+$KL (VRF); bases frozen \\
\midrule
Response embedding & all-MiniLM-L6-v2 (384-dim) \\
Random seeds & 20 (42--61) \\
Bandit epochs & 10 \\
Total interactions (10 epochs) & 36,560 \\
\bottomrule
\end{tabular}
\end{table}

\section{Results}
\label{sec:results}

We present results as controlled comparisons, each isolating one mechanism: (1)~online updates vs.\ static models, (2)~exploration vs.\ greedy, (3)~directional forgetting vs.\ uniform or no forgetting, (4)~error-conditional vs.\ unconditional injection.

\textbf{Baseline: stationarity.} Without drift, \ours{} (0.681) and Greedy (0.684) perform similarly ($p{=}0.066$; \Cref{tab:stationary}, \Cref{app:stationary}), confirming that \ours{} incurs minimal cost under stable preferences.

We now evaluate under the flip protocol defined in \Cref{sec:config}. Each subsection isolates the next mechanism in the cascade.

\subsection{Online Updates vs.\ Static Models}
\label{sec:flip}

We invert rewards for a fraction $f \in \{0.25, 0.5, 0.75\}$ of users after pretraining ($r \to r_{\max} + r_{\min} - r$), making the model maximally wrong in its most confident directions (20 seeds). \ours{} retains its advantage under gradual and decelerating protocols (\Cref{tab:gradual_drift}, \Cref{app:gradual_drift}).

\begin{table}[t]
\centering
\caption{PRISM flip: average reward at epoch 10 (20 seeds). \ours{} leads all baselines across flip severities, with the advantage over NeuralUCB-A widening from marginal ($+$0.2 pts) at $f{=}0.25$ to $+$2.5--2.7 pts at $f{\geq}0.5$. Stationary results (flip${=}0.0$) in \Cref{app:stationary}.}
\label{tab:flip}
\small
\setlength{\tabcolsep}{3.5pt}
\begin{tabular}{@{}lccccccccc@{}}
\toprule
\textbf{Flip} & \textbf{Oracle} & \textbf{\ours{}} & \textbf{NUCB-A} & \textbf{NUCB-F} & \textbf{NeuralTS} & \textbf{Greedy} & \textbf{No-Onl.} & \textbf{Random} \\
 & & $\gamma{=}4$ & $\eta{=}0$ & & & & & \\
\midrule
0.25 & .875 & \textbf{.651} & .649 & .639 & .615 & .631 & .604 & .548 \\
0.50 & .875 & \textbf{.641} & .616 & .610 & .593 & .601 & .541 & .542 \\
0.75 & .875 & \textbf{.667} & .640 & .632 & .598 & .621 & .480 & .537 \\
\bottomrule
\end{tabular}
\end{table}

The same method ranking holds in the controlled simulation (\Cref{app:sim_flip}).

\textbf{(1) Online updates are essential.} No-Online collapses to 0.480 at flip 0.75, well below Random (0.537): a model that cannot update becomes actively harmful under drift. This is the single largest effect and the prerequisite for everything that follows. Offline reward models (MRM, VRF) are essentially No-Online baselines.

\textbf{(2) Exploration beats greedy under drift.} At flip 0.5, \ours{} leads Greedy by $+$4.0 points ($p{<}0.001$). When user preferences shift, greedy exploitation of stale predictions is costly; any form of exploration helps. Under stationarity, \ours{} is within 0.3 points of Greedy ($p{=}0.066$; \Cref{app:stationary}), confirming that exploration activates only when needed. At flip 0.5, \ours{} closes 29.7\% of the Random-to-Oracle gap versus Greedy's 17.7\%.

\textbf{(3) Directional forgetting beats uniform and no forgetting.} The $+$2.5-point gap from NeuralUCB-A ($\eta{=}0$) to \ours{} ($\eta{=}0.01$) isolates the value of the bidirectional Gram update beyond error-adaptive exploration alone. This gap \textbf{widens with flip severity} from 0.25 to 0.75, as predicted by the regret decomposition (\Cref{cor:nonstationary_regret}, \Cref{app:bayesian_regret}; instantiated in \Cref{cor:concrete_scaling}, \Cref{app:adaptive_regret}): more users experience errors that trigger injection, growing uncertainty in precisely the stale directions. Epoch-1 significance tests (\Cref{tab:epoch1_significance}, \Cref{app:epoch_sweep}) confirm this is genuine learning, not noise.

\begin{figure}[ht!]
\centering
\includegraphics[width=0.85\textwidth]{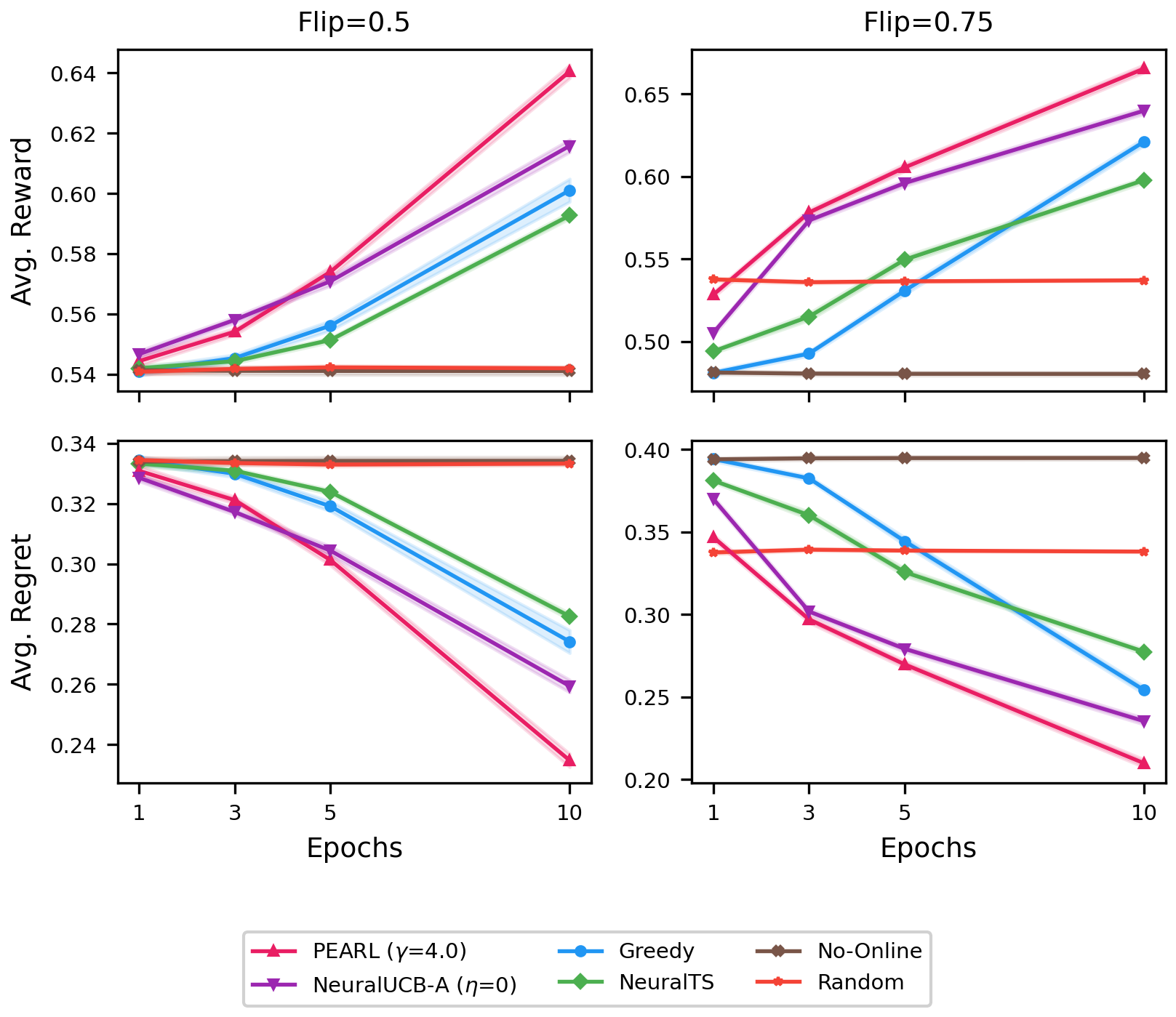}
\caption{Learning dynamics under preference flip (20 seeds, shaded regions show $\pm$1 standard error). Top row: average reward; bottom row: average regret. At flip 0.5 (left), \ours{} ($\gamma{=}4.0$) separates from Greedy mid-training. At flip 0.75 (right), the gap widens further, No-Online collapses below Random, and the adaptive advantage is clear by epoch 3.}
\label{fig:flip_epoch_sweep}
\end{figure}

Learning dynamics (\Cref{fig:flip_epoch_sweep}, \Cref{app:dynamics_figures}) show \ours{}'s advantage emerging mid-training, with drift recovery trajectories confirming accelerated recovery for flipped contexts while non-flipped contexts remain stable. The preference-error heatmap (\Cref{fig:pref_heatmap}, \Cref{app:pref_validity}) shows different dimensions peaking and decaying at different rates, confirming dimension-specific re-exploration.

\textbf{Gradual drift.} The abrupt flip is an adversarial stress test. We also evaluate \emph{decelerating drift} (rapid shift over 3 epochs, then stable). \ours{} leads Greedy by $+$3.6 points (\Cref{tab:gradual_drift}, \Cref{app:gradual_drift}), confirming that the mechanism is not dependent on abrupt changes.

\subsection{Baselines and Ablations}
\label{sec:baselines}

We compare \ours{} against non-stationary bandits (SW-UCB, $W \in \{200, 500, 1000\}$), alternative reward architectures (URM, MRM, VRF), and ablations (flip 0.5, epoch 10, 20 seeds). Full results in \Cref{tab:comparison} (\Cref{app:full_comparison}); \Cref{tab:ablation_cascade} summarizes each mechanism's additive contribution.

\begin{table}[h]
\centering
\caption{Ablation cascade on PRISM (flip 0.5, epoch 10, 20 seeds). Each row adds one mechanism; the $\Delta$ column shows its marginal contribution.}
\label{tab:ablation_cascade}
\small
\begin{tabular}{@{}llcc@{}}
\toprule
\textbf{Configuration} & \textbf{What it adds} & \textbf{Reward} & $\boldsymbol{\Delta}$ \\
\midrule
No-Online & (static model) & .541 & \\
Greedy & $+$ online updates & .601 & $+$.060 \\
NeuralUCB-F & $+$ fixed exploration & .610 & $+$.009 \\
NeuralUCB-A & $+$ error-adaptive expl. & .616 & $+$.006 \\
\ours{} & $+$ bidirectional Gram & \textbf{.641} & $+$.025 \\
\bottomrule
\end{tabular}
\end{table}

The small $\Delta$ from NeuralUCB-F to NeuralUCB-A ($+$0.6 pts) is expected: error-adaptive weighting enables the bidirectional Gram to target the right directions, so its value appears in the next row ($+$2.5 pts). Under stationarity it matches greedy; under drift it gates directional forgetting.

\textbf{Uniform forgetting methods.} The best SW-UCB ($W{=}1000$) achieves only 0.560, barely above Random (0.542); even NeuralUCB-A ($\eta{=}0$) outperforms it by 5.6 points. Discounted NeuralUCB either matches NeuralUCB-A ($\gamma_{\text{gram}}{=}0.999$) or over-forgets ($\gamma_{\text{gram}}{=}0.99$, \Cref{tab:discounted_ucb}, \Cref{app:nonstationary_baselines}); Restarting NeuralUCB still trails \ours{} by 2 points (\Cref{app:nonstationary_baselines}). All three forget uniformly, unable to distinguish stale directions from valid ones.

\textbf{Alternative reward architectures.} MRM (0.554), URM (0.548), and VRF (0.544) barely beat Random. URM, MRM, and VRF were not designed for bandits; we include them to test whether their uncertainty estimates can drive exploration under drift. Adding Gram-matrix exploration on top would not resolve the mismatch: URM's heteroscedastic $\sigma^2$ captures aleatoric rather than epistemic uncertainty, MRM's bases freeze after pretraining so a Gram matrix over stale features is uninformative, and VRF already provides a variational uncertainty source with no principled way to layer a second estimate on top. The common failure is that none track epistemic uncertainty over online-updated features. Full results in \Cref{tab:comparison} (\Cref{app:full_comparison}).

We verify additional design choices: conditional injection outperforms unconditional always-inject (\Cref{app:always_inject}), results are insensitive to $\eta$ (\Cref{tab:eta_combined}, \Cref{app:eta_ablation}), and method rankings hold across pretrain fractions (\Cref{app:pretrain_fraction}), filtering thresholds (\Cref{app:filtering_sensitivity}), embedding dimensions (\Cref{app:embed_ablation}), and $\lambda$ values (\Cref{app:lambda_ablation}).

\section{Stationary Performance on PRISM (Flip 0.0)}
\label{app:stationary}

Under stationarity on PRISM (no preference flip), exploration provides no benefit. \Cref{tab:stationary} reports epoch-10 results. \ours{} (0.681) is within 0.3 points of Greedy (0.684); a paired $t$-test across 20 seeds yields $p{=}0.066$ (not significant at $\alpha{=}0.05$). \ours{}'s exploration mechanism incurs negligible cost when preferences are stable, while providing significant gains under partial stationarity (\Cref{tab:flip}).

\begin{table}[ht!]
\centering
\caption{Stationary performance (flip 0.0, epoch 10, 20 seeds, paired $t$-test vs.\ Greedy). \ours{}'s exploration does not significantly hurt performance: the 95\% CI for \ours{}$-$Greedy spans zero.}
\label{tab:stationary}
\small
\begin{tabular}{lcccl}
\toprule
\textbf{Method} & \textbf{Reward} & \textbf{$\Delta$ vs.\ Greedy} & \textbf{95\% CI} & \textbf{$p$} \\
\midrule
Greedy & \textbf{.684} & --- & --- & --- \\
\ours{} ($\gamma{=}4$, $\eta{=}0.01$) & .681 & $-$.003 & [$-$.005, $+$.000] & .066 \\
No-Online & .664 & $-$.020 & [$-$.021, $-$.018] & $<$.001 \\
Random & .554 & $-$.130 & [$-$.131, $-$.128] & $<$.001 \\
\bottomrule
\end{tabular}
\end{table}

\section{Synthetic Simulation}
\label{app:sim_flip}

To complement the real-data evaluation on PRISM, we construct a minimal synthetic environment that isolates the core mechanism: when a pretrained model becomes confidently wrong for every user, can error-adaptive exploration recover faster than fixed-schedule alternatives?

\paragraph{Environment.} Ten users have 5-dimensional preference vectors drawn from $\text{Beta}(2,2)$, representing latent taste along interpretable axes (analogous to formality, detail, creativity, technicality, and conciseness in an LLM setting). A fixed pool of $K{=}10$ candidates, each characterized by a 5-dim feature vector with one dominant dimension, is available at every interaction. At each step, the bandit observes the user's preference features and selects one candidate to recommend; the reward is the inner product between the user's preference vector and the selected candidate's features (clipped to $[0,1]$), plus Gaussian noise ($\sigma{=}0.05$). Only the reward for the selected candidate is observed, matching the partial-feedback structure of a real deployment.

\paragraph{Pretraining and preference shift.} We pretrain on 500 random interactions using Bradley--Terry pairwise loss, giving the model a confident prior about each user's preferred candidates. We then apply a complete preference inversion ($f{=}1.0$): all user preference vectors are negated, so previously preferred candidates become the worst choices and vice versa. This is deliberately more extreme than the partial flips in PRISM ($f{=}0.5$), serving as a stress test that makes the pretrained model maximally stale for every user simultaneously.

\paragraph{Why this setting is informative.} The fixed candidate pool prevents Greedy from passively discovering better alternatives. In PRISM, different LLM responses appear each interaction, offering some natural diversity even without deliberate exploration. Here, Greedy repeatedly selects its pretrained favorite, observes low rewards, but gains no gradient signal about the other 9 candidates. It must actively try new arms to learn about them.

\paragraph{Online learning protocol.} Each epoch reshuffles the same 50 bandit interactions (approximately 5 per user). The model updates once per epoch via MSE on accumulated observations. Between updates, the bidirectional Gram injection fires on each step where the prediction error exceeds the user's running average, inflating uncertainty along the feature direction of the surprising observation and directing future exploration toward stale user--arm pairs.

\begin{table}[ht!]
\centering
\caption{Synthetic simulation under complete preference inversion ($f{=}1.0$), epoch 3. \ours{} recovers fastest from worst-case staleness, leading all baselines. Config: 10 users, $K{=}10$ fixed arms, $d{=}5$, $\gamma{=}100$, $\eta{=}0.01$.}
\label{tab:sim_comparison}
\small
\begin{tabular}{@{}lcc@{}}
\toprule
\textbf{Method} & \textbf{Reward} & \textbf{vs.\ \ours{}} \\
\midrule
\ours{} ($\gamma{=}100$, $\eta{=}0.01$) & \textbf{.848} & --- \\
NeuralUCB ($\alpha{=}20$) & .795 & $-$.053 \\
Random & .776 & $-$.072 \\
SW-UCB ($W{=}50$) & .754 & $-$.094 \\
NeuralTS & .734 & $-$.114 \\
Greedy & .711 & $-$.137 \\
\bottomrule
\end{tabular}
\end{table}

\paragraph{Results.} \Cref{tab:sim_comparison} reports epoch-3 performance, after the model has had two online updates to begin adapting. \ours{} leads Greedy by $+$13.7 points and NeuralUCB by $+$5.3 points. The method ranking mirrors the real-data evaluation: \ours{} $>$ NeuralUCB $>$ Random $>$ SW-UCB $>$ NeuralTS $>$ Greedy.

The gap between \ours{} and NeuralUCB is particularly informative: both methods use UCB exploration, but \ours{} concentrates its exploration on user--arm pairs where the model is confidently wrong (via bidirectional Gram injection), while NeuralUCB explores uniformly based on the standard shrink-only Gram uncertainty. This directional targeting allows \ours{} to discover the new optimal arms faster.

Greedy remains near 0.71 throughout. It keeps selecting the now-worst arm and, despite observing poor rewards, cannot generalize to untried alternatives because the fixed candidate pool offers no passive diversity. Random (0.776) outperforms Greedy by simply sampling all arms uniformly, confirming that the pretrained prior actively hurts when stale.

\section{Learning Dynamics and Drift Recovery Figures}
\label{app:dynamics_figures}

We visualize learning trajectories and drift recovery to complement the epoch-10 summary statistics in \Cref{tab:flip}.

\begin{figure}[ht!]
\centering
\includegraphics[width=\textwidth]{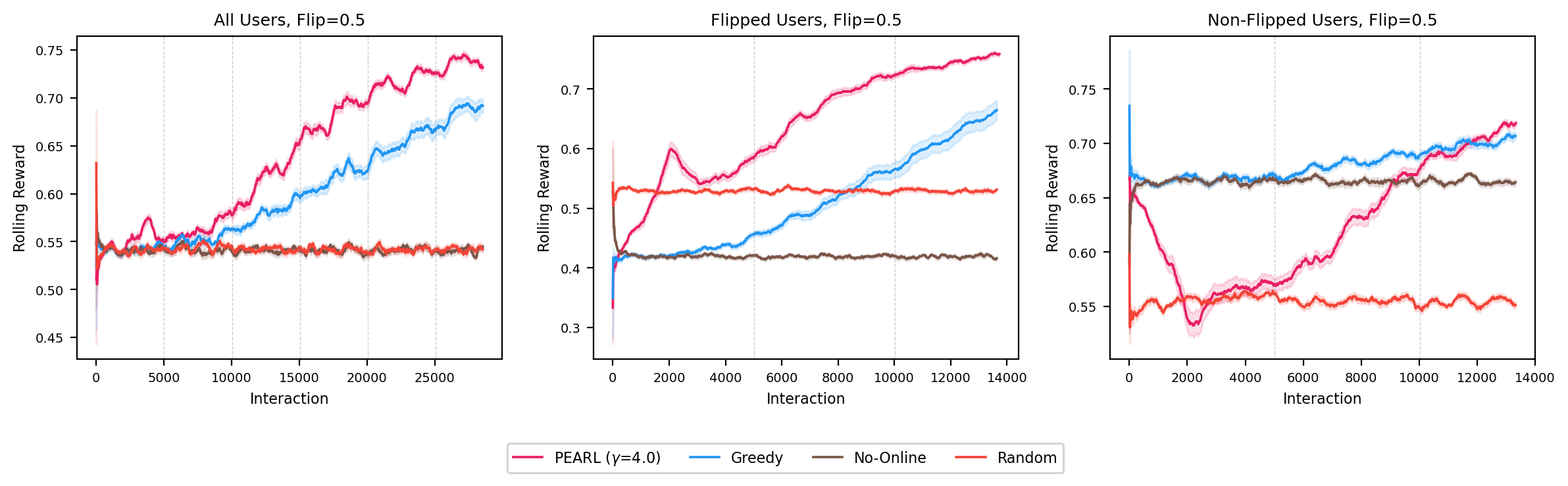}
\caption{Drift recovery trajectories under flip $f{=}0.5$ (20 seeds, rolling window of 500 interactions). Left: all users. Center: flipped users only. Right: non-flipped users only. Flipped users start near 0.42 and recover to 0.75; non-flipped users maintain stable performance. \ours{} ($\gamma{=}4.0$) separates from Greedy mid-recovery. No-Online flatlines in all conditions.}
\label{fig:drift_recovery}
\end{figure}

For non-flipped users (\Cref{fig:drift_recovery}, right), the methods perform similarly, while for flipped users (\Cref{fig:drift_recovery}, center) \ours{} leads by $+$5 pts, yielding a net advantage in the aggregate (\Cref{tab:flip}). The error-adaptive mechanism concentrates exploration on users whose predictions are poor, reducing unnecessary exploration on well-modeled users. The primary advantage is speed of recovery: \ours{} reaches good performance earlier, which matters in settings where each interaction is costly.

\section{Pretrain Fraction Ablation}
\label{app:pretrain_fraction}

We vary the fraction of data used for pretraining ($\{25\%, 50\%, 75\%\}$) to test whether \ours{}'s advantage depends on pretrain volume (\Cref{tab:pretrain_fraction}). Larger pretrain fractions create stronger stale priors, amplifying the benefit of error-adaptive exploration.

\begin{table}[ht!]
\centering
\caption{Pretrain fraction ablation (epoch 10, 20 seeds). \ours{}'s advantage over Greedy (A$-$G) is substantial across conditions. The gap grows with pretrain fraction (the ``stale prior'' effect): stronger pretraining creates more misleading priors, amplifying the benefit of error-adaptive exploration and bidirectional Gram updates.}
\label{tab:pretrain_fraction}
\small
\begin{tabular}{llccc}
\toprule
\textbf{Pretrain} & \textbf{Flip} & \textbf{Greedy} & \textbf{\ours{} ($\gamma{=}4.0$)} & \textbf{A$-$G} \\
\midrule
25\% & 0.5 & .634 & .664 & $+$.030 \\
50\% (main) & 0.5 & .601 & .641 & $+$.040 \\
75\% & 0.5 & .592 & .627 & $+$.035 \\
\bottomrule
\end{tabular}
\end{table}

\section{Filtering Sensitivity}
\label{app:filtering_sensitivity}

Our main experiments filter to interactions with reward spread ${>}0.3$ (18.5\% of data). We relax this threshold to ${>}0.1$ and ${>}0.2$ to verify that method rankings are not an artifact of this filtering choice (\Cref{tab:filtering}).

\begin{table}[ht!]
\centering
\caption{Filtering sensitivity under flip 0.5 (20 seeds). Method rankings and the adaptive-greedy gap are preserved across filtering thresholds.}
\label{tab:filtering}
\small
\begin{tabular}{lccc}
\toprule
\textbf{Method} & \textbf{Spread $>$0.1} & \textbf{Spread $>$0.2} & \textbf{Spread $>$0.3} \\
\midrule
Adaptive ($\gamma{=}4.0$) & .649 $\pm$ .011 & .639 $\pm$ .019 & .641 $\pm$ .015 \\
Greedy & .625 $\pm$ .008 & .618 $\pm$ .014 & .613 $\pm$ .017 \\
\midrule
Gap (A$-$G) & $+$.024 & $+$.021 & $+$.028 \\
\bottomrule
\end{tabular}
\end{table}

\section{Personalization Ablation ($d_u{=}0$)}
\label{app:nopers}

We ablate user features by setting $d_u{=}0$ (removing both stated preferences and learned embeddings) while keeping $\gamma{=}4.0$ (error-adaptive exploration). This isolates the contribution of user features from the exploration mechanism (\Cref{tab:nopers}).

\begin{table}[ht!]
\centering
\caption{No-Pers ablation: \ours{} $\gamma{=}4.0$ with and without user features (20 seeds, epoch 10). User features help at partial flip (0.25--0.75) by identifying which users need re-exploration.}
\label{tab:nopers}
\small
\begin{tabular}{lccc}
\toprule
\textbf{Method} & \textbf{Flip 0.25} & \textbf{Flip 0.5} & \textbf{Flip 0.75} \\
\midrule
\ours{} $\gamma{=}4.0$ ($d_u{=}40$) & .651 & \textbf{.641} & \textbf{.667} \\
\ours{} $\gamma{=}4.0$ ($d_u{=}0$)  & \textbf{.642} & .612 & .646 \\
Greedy                               & .631 & .601 & .621 \\
\midrule
$d_u{=}40$ vs.\ $d_u{=}0$ & $+$.009 & $+$.029 & $+$.021 \\
\bottomrule
\end{tabular}
\end{table}

At partial flip (0.5, 0.75), user embeddings provide $+$1.9 to $+$2.9 points by helping the model identify which users have shifted preferences. Under mild flip (0.25), $d_u{=}40$ leads by 0.9 points. User features are most valuable when preference shift is heterogeneous across the population, enabling the model to distinguish which users need re-exploration.

\section{Candidate Diversity Stratification}
\label{app:diversity}

A natural concern is whether \ours{}'s advantage depends on high candidate diversity (as in the 21-LLM pool). To test this, we compute the average pairwise cosine distance between candidate SBERT embeddings for each user's interactions and split users into diversity terciles: low (${<}0.35$), mid (0.35--0.47), and high (${\geq}0.47$), with ${\sim}$450 users each.

\begin{table}[ht!]
\centering
\caption{Candidate diversity stratification under flip 0.5 (20 seeds, epoch 10). \ours{}'s advantage is present across all diversity levels but amplified by higher candidate diversity. Low-diversity interactions simulate the reduced distinctiveness of single-model Best-of-N candidates. Tercile averages differ from the population average in \Cref{tab:flip} due to unequal interaction counts across groups.}
\label{tab:diversity}
\small
\begin{tabular}{lccc}
\toprule
\textbf{Method} & \textbf{Low div.} & \textbf{Mid div.} & \textbf{High div.} \\
\midrule
\ours{} ($\gamma{=}4.0$) & .635 & .617 & .601 \\
Greedy & .625 & .600 & .578 \\
\midrule
Gap & $+$.010*** & $+$.018*** & $+$.023*** \\
\bottomrule
\end{tabular}
\end{table}

Under partial stationarity, \ours{} significantly outperforms Greedy across all diversity terciles ($p{<}0.01$). The advantage is smallest for low-diversity candidates ($+$1.0 pts at flip 0.5) and largest for high-diversity candidates ($+$2.3 pts). Low-diversity interactions approximate the reduced candidate distinctiveness of single-model Best-of-N sampling (where candidates come from temperature variation of one model rather than architecturally distinct LLMs). \ours{}'s advantage persists in this regime, confirming that the mechanism does not depend on extreme candidate diversity, though diversity amplifies the benefit by providing more room for targeted selection.

\section{Epoch Sweep Details Under Preference Flips}
\label{app:epoch_sweep}

\Cref{tab:epoch_sweep_flip} reports reward at each epoch checkpoint, showing the full learning trajectory rather than only the final epoch-10 values from \Cref{tab:flip}.

\begin{table}[ht!]
\centering
\caption{Average reward at each epoch checkpoint under preference flips (20 seeds). By epoch~10, \ours{} leads at flip 0.5. No-Online is flat across epochs; at flip ${\geq}0.5$ it falls below Random.}
\label{tab:epoch_sweep_flip}
\small
\begin{tabular}{cl cccc}
\toprule
\textbf{Flip} & \textbf{Method} & \textbf{Epoch 1} & \textbf{Epoch 3} & \textbf{Epoch 5} & \textbf{Epoch 10} \\
\midrule
0.50 & \ours{} ($\gamma{=}4.0$)      & .544 & .554 & .574 & \textbf{.641} \\
     & NeuralUCB-A ($\eta{=}0$)      & .547 & .558 & .571 & .616 \\
     & Greedy                        & .541 & .545 & .556 & .601 \\
     & No-Online                     & .541 & .541 & .541 & .541 \\
     & Random                        & .541 & .542 & .542 & .542 \\
\bottomrule
\end{tabular}
\end{table}

\paragraph{Epoch-1 significance tests.} To verify that multi-epoch results reflect genuine cumulative learning rather than replay inflation, we test whether \ours{} already leads at epoch~1 (a single pass through each user's interactions). \Cref{tab:epoch1_significance} reports paired $t$-tests (20 seeds, BCa bootstrap 95\% CI, 10{,}000 resamples).

\begin{table}[ht!]
\centering
\caption{Epoch-1 significance tests (20 seeds, paired $t$-test). \ours{}'s epoch-1 advantage is significant only at flip${\geq}0.75$; at flip~0.5 the $+$0.4-point lead is not significant ($p{=}0.118$). At flip~0.25, Greedy leads at epoch~1 ($p{<}0.001$). This confirms that \ours{}'s value emerges through cumulative online learning, not from a single-pass advantage.}
\label{tab:epoch1_significance}
\small
\setlength{\tabcolsep}{3.5pt}
\begin{tabular}{@{}llcccc@{}}
\toprule
\textbf{Flip} & \textbf{Comparison} & \textbf{$\Delta$} & \textbf{95\% CI} & \textbf{$p$} & \textbf{Sig.} \\
\midrule
0.25 & \ours{}$-$Greedy & $-$.013 & [$-$.019, $-$.008] & ${<}.001$ & Greedy$\uparrow$ \\
0.50 & \ours{}$-$Greedy & $+$.004 & [$-$.001, $+$.009] & .118 & n.s. \\
0.75 & \ours{}$-$Greedy & $+$.047 & [$+$.039, $+$.055] & ${<}.001$ & \ours{}$\uparrow$ \\
\bottomrule
\end{tabular}
\end{table}

At epoch~1, all methods see each user only once; the exploration mechanism has minimal opportunity to differentiate. At flip~0.5, \ours{} and Greedy are statistically tied. At flip~0.75, however, \ours{} already leads by $+$4.7 points ($p{<}0.001$, $d{=}4.92$): with 75\% of users flipped, the error tracker immediately activates, and the first-pass exploration bonus is already directionally useful. The advantage then widens through subsequent epochs (\Cref{tab:epoch_sweep_flip}), confirming that the cumulative epoch-10 results reflect genuine multi-pass learning.

\section{Reward Model Calibration Analysis}
\label{app:calibration}

To assess whether the model's predicted mean rewards are well-calibrated, we compute the Expected Calibration Error (ECE) and reliability diagrams for each method across flip conditions. ECE measures the weighted average absolute difference between predicted and observed rewards across 10 prediction bins, evaluated on the last 10\% of interactions (approximately epoch 10).

\begin{figure}[ht!]
\centering
\includegraphics[width=0.8\textwidth]{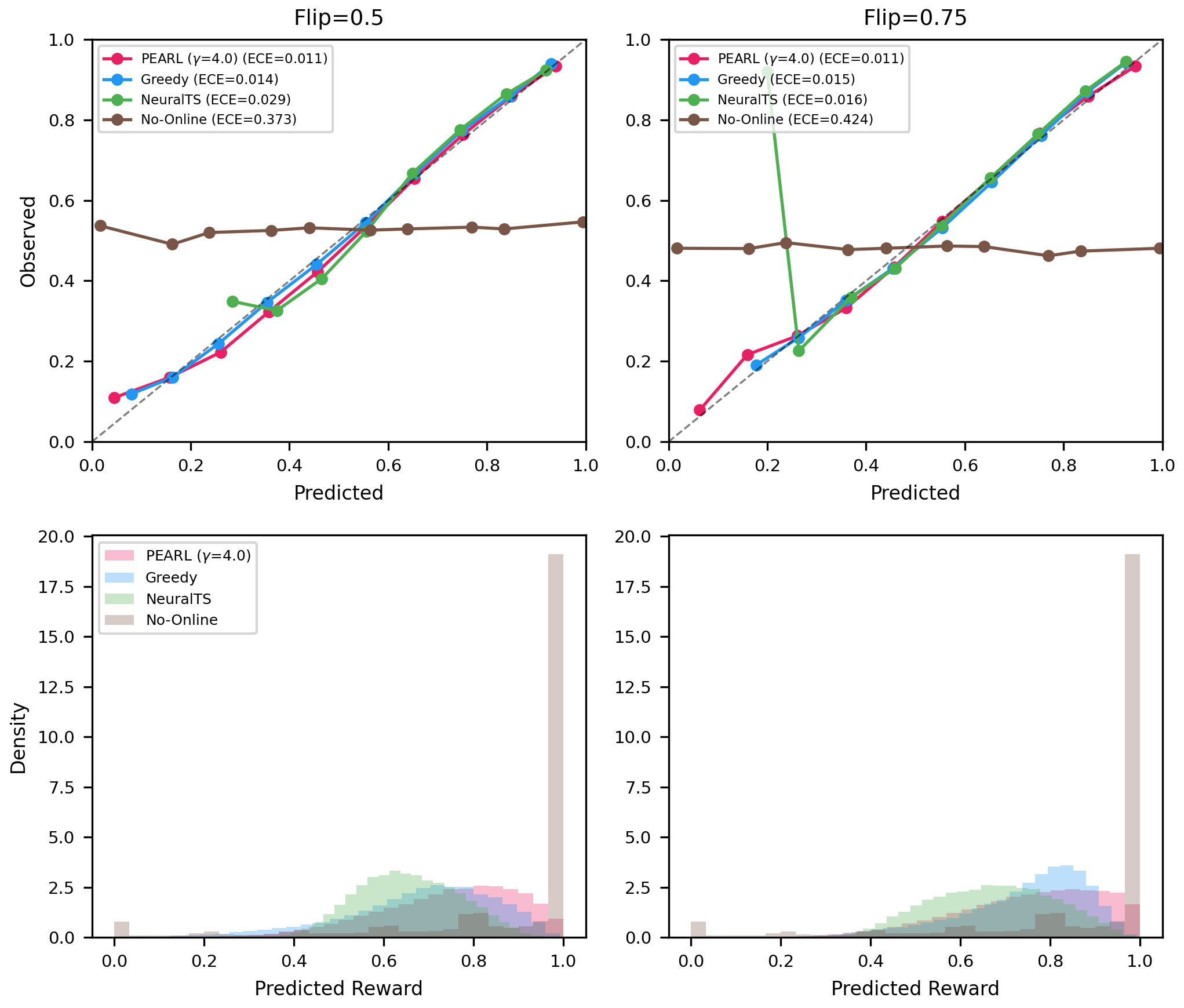}
\caption{Reward model calibration (20 seeds). Top row: reliability diagrams showing predicted vs.\ observed reward. Bottom row: prediction distributions. All online-adapted methods (\ours{}, Greedy, NeuralTS) achieve excellent calibration (ECE $<$ 0.03) across both flip conditions. No-Online's calibration degrades sharply (ECE 0.37 at flip 0.5 to 0.42 at flip 0.75) as its stale predictions diverge from shifted preferences.}
\label{fig:calibration}
\end{figure}

\Cref{fig:calibration} separates the methods into two groups. All online-adapted methods, \ours{} (ECE 0.008--0.011), Greedy (0.011--0.014), and NeuralTS (0.008--0.029), maintain excellent calibration across all flip conditions. Their reliability curves hug the diagonal, confirming that online learning successfully recalibrates the model even under severe preference drift. By contrast, No-Online's calibration error grows substantially with flip severity; its reliability curve flattens near 0.55 observed reward regardless of predicted value, reflecting that its stale predictions are uncorrelated with post-flip rewards. The prediction distributions (bottom row) show that No-Online's predictions concentrate near 0.9--1.0 regardless of flip severity, while online methods maintain a spread that reflects actual reward variability.

\paragraph{Implications for UCB exploration.} The low ECE of \ours{} confirms that the Gram-matrix uncertainty estimates remain meaningful under partial stationarity. Since UCB exploration relies on calibrated uncertainty to direct exploration toward high-potential arms, this calibration is a necessary condition for the error-adaptive and bidirectional Gram mechanisms to function correctly.

\paragraph{Gram uncertainty dynamics.} Under mild drift, the Gram uncertainty $\sigma_k$ spikes early as the bidirectional update detects prediction errors, then decays as the model recovers. Higher flip fractions produce larger initial spikes. This spike-then-decay pattern mirrors the EMA error tracker dynamics and confirms that the bidirectional Gram update responds proportionally to the severity of preference drift.

\section{IQM and Bootstrap Confidence Intervals}
\label{app:iqm}

Following \citet{agarwal2021deep}, we report the Interquartile Mean (IQM) with 95\% BCa bootstrap confidence intervals (10,000 resamples) as a robust alternative to the mean. IQM averages only the middle 50\% of runs, reducing sensitivity to outliers. \Cref{tab:iqm} confirms that all method rankings from \Cref{tab:flip} are preserved under this robust aggregate, and the narrow CIs reflect low variance across seeds.

\begin{table}[htbp]
\centering
\scriptsize
\caption{Interquartile Mean (IQM) of average reward at epoch 10 with 95\% BCa bootstrap confidence intervals (10,000 resamples), following \citet{agarwal2021deep}. IQM averages the middle 50\% of runs, reducing sensitivity to outliers.}
\label{tab:iqm}
\setlength{\tabcolsep}{4pt}
\begin{tabular}{lccc}
\toprule
Method & Flip 0.25 & Flip 0.5 & Flip 0.75 \\
\midrule
\ours{} ($\gamma{=}4$) & .652 {\tiny[.647,.655]} & .642 {\tiny[.638,.646]} & .667 {\tiny[.660,.671]} \\
NUCB-A ($\eta{=}0$) & .648 {\tiny[.645,.653]} & .615 {\tiny[.610,.620]} & .640 {\tiny[.635,.645]} \\
NUCB-F & .639 {\tiny[.635,.643]} & .609 {\tiny[.604,.615]} & .632 {\tiny[.626,.637]} \\
NeuralTS & .615 {\tiny[.613,.616]} & .592 {\tiny[.590,.595]} & .596 {\tiny[.593,.602]} \\
Greedy & .631 {\tiny[.628,.636]} & .600 {\tiny[.591,.609]} & .623 {\tiny[.618,.626]} \\
No-Online & .604 {\tiny[.601,.607]} & .542 {\tiny[.539,.545]} & .480 {\tiny[.478,.483]} \\
Random & .548 {\tiny[.547,.549]} & .542 {\tiny[.540,.543]} & .537 {\tiny[.536,.538]} \\
\bottomrule
\end{tabular}
\end{table}

\section{Stratified Analysis by User Divergence}
\label{app:stratified}

To test whether \ours{}'s advantage varies with user characteristics, we stratify users by \emph{preference divergence}, defined as how difficult a user's preferences are to predict. We measure divergence as each user's average EMA prediction error ($\varepsilon_u$) in the stationary condition (flip 0), then split users into terciles (Low, Medium, High).

\begin{figure}[ht!]
\centering
\includegraphics[width=\textwidth]{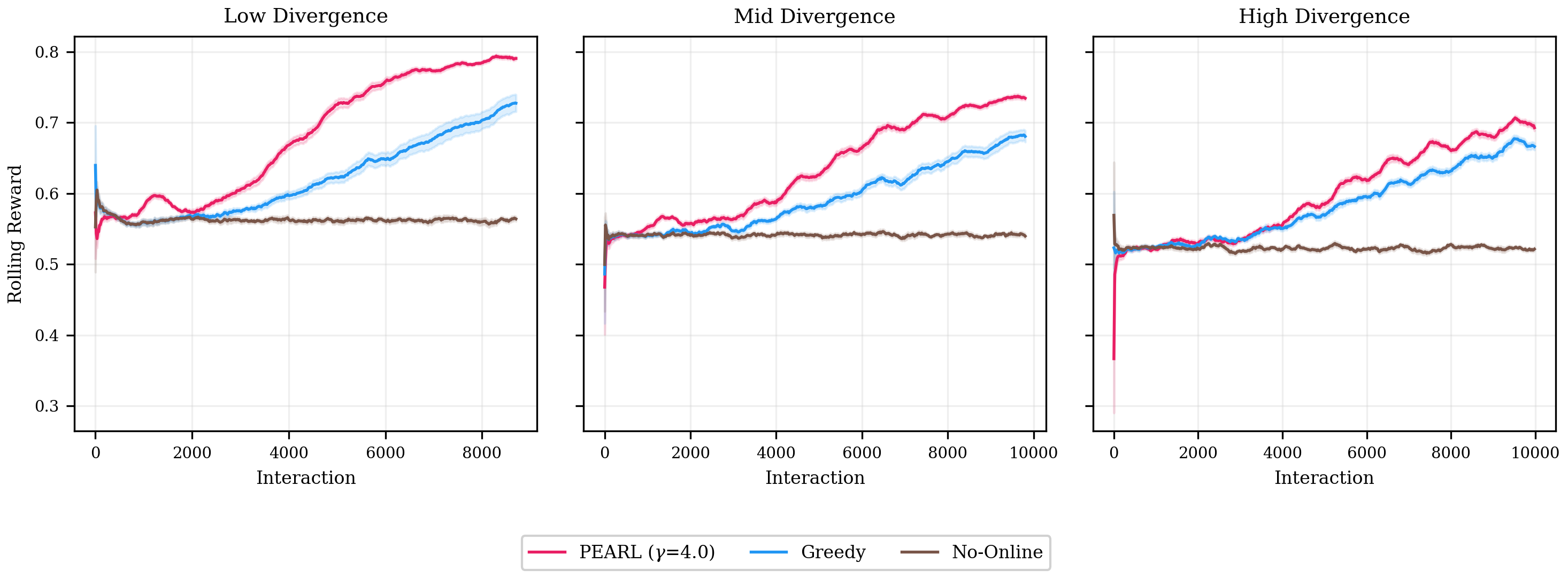}
\caption{Performance stratified by user preference divergence (20 seeds, tercile split by EMA prediction error). At flip 0.5, \ours{}'s advantage over Greedy is largest for low-divergence users (+3.9 pts), who were well-modeled before the flip and thus most disrupted by it. No-Online collapses across all groups under flip.}
\label{fig:stratified_divergence}
\end{figure}

\Cref{fig:stratified_divergence} shows that \ours{}'s gains depend on pre-flip model quality. Under flip 0.5, \ours{}'s advantage over Greedy is largest for \textbf{low-divergence users} (+3.9 pts) and moderate for medium-divergence users (+1.9 pts), but negligible for high-divergence users. This is because low-divergence users were well-modeled before the flip: their reward predictions were accurate, so the flip creates a large prediction error spike that triggers \ours{}'s error-adaptive exploration. High-divergence users were poorly modeled even before the flip, so the flip changes little about their error profile. No-Online collapses across all groups, confirming that the failure is universal rather than confined to a subpopulation.

\section{Full Baseline Comparisons (PRISM)}
\label{app:full_comparison}

\begin{table}[h]
\centering
\caption{PRISM comparison under flip 0.5 at epoch 10 (20 seeds). \ours{} dominates SW-UCB by 8+ points and alternative reward models by 9--10 points. Setting $\eta{=}0$ (NeuralUCB-A, no bidirectional injection) ablates the Gram update, costing 2.5 points.}
\label{tab:comparison}
\small
\begin{tabular}{lcc}
\toprule
\textbf{Method} & \textbf{Reward} & \textbf{vs.\ \ours{}} \\
\midrule
\ours{} ($\gamma{=}4$, $\eta{=}0.01$) & \textbf{.641} & --- \\
NeuralUCB-A ($\eta{=}0$) & .616 & $-$.025 \\
NeuralUCB-F & .610 & $-$.031 \\
Greedy & .601 & $-$.040 \\
NeuralTS & .593 & $-$.048 \\
\midrule
SW-UCB ($W{=}1000$) & .560 & $-$.081 \\
SW-UCB ($W{=}500$) & .547 & $-$.094 \\
SW-UCB ($W{=}200$) & .544 & $-$.097 \\
\midrule
MRM (adaptive) & .554 & $-$.087 \\
URM (adaptive) & .548 & $-$.093 \\
VRF (adaptive) & .544 & $-$.097 \\
\midrule
Random & .542 & $-$.099 \\
\bottomrule
\end{tabular}
\end{table}

\Cref{tab:comparison} presents the full baseline comparison on PRISM under moderate drift (flip 0.5). First, \ours{} leads NeuralUCB-A by 2.5 points, isolating the contribution of the bidirectional Gram update (NeuralUCB-A is identical to \ours{} but with $\eta{=}0$, disabling injection). Second, all three SW-UCB window sizes underperform by 8--10 points, confirming that indiscriminate forgetting is worse than no forgetting at all. Third, recent reward architectures (MRM, URM, VRF) cluster near Random, suggesting that richer reward modeling alone does not compensate for the lack of online adaptation.

\section{Injection Rate ($\eta$) Sensitivity}
\label{app:eta_ablation}

The injection rate $\eta$ controls how aggressively the bidirectional Gram update grows uncertainty when the model is surprisingly wrong (\cref{eq:injection}). Setting $\eta{=}0$ recovers standard NeuralUCB with error-adaptive exploration (no bidirectional injection). \Cref{tab:eta_combined} consolidates all $\eta$-related ablations under flip 0.5 (20 seeds, epoch 10).

\begin{table}[ht!]
\centering
\caption{Injection rate ablations (20 seeds, epoch 10, flip 0.5). All rows are \ours{} variants; $\eta{=}0$ recovers NeuralUCB-A (no injection). The bidirectional update provides $+$2.5 pts over $\eta{=}0$ and is robust across an order of magnitude. PEARL-auto scales $\eta_{\text{eff}} = \eta_{\text{base}} \cdot \bar{\varepsilon}/0.5$. Always-inject fires unconditionally, without the error condition.}
\label{tab:eta_combined}
\small
\begin{tabular}{lc}
\toprule
\textbf{Variant} & \textbf{Reward} \\
\midrule
$\eta{=}0$ (NeuralUCB-A) & .616 \\
$\eta{=}0.001$ & .645 \\
$\eta{=}0.01$ (default) & \textbf{.641} \\
$\eta{=}0.1$ & .634 \\
\midrule
Auto $\eta_{\text{base}}{=}0.01$ & \textbf{.647} \\
Auto $\eta_{\text{base}}{=}0.05$ & .635 \\
Auto $\eta_{\text{base}}{=}0.1$ & .635 \\
\midrule
Always-inject ($\eta{=}0.01$) & .637 \\
\bottomrule
\end{tabular}
\end{table}

Performance is robust across an order of magnitude of $\eta$: the spread is at most 2.9 pts (fixed) and 1.2 pts (auto). $\eta_{\text{base}}{=}0.01$ is optimal, confirming that the adaptive mechanism makes this choice non-critical.

\paragraph{Always-inject baseline.}
\label{app:always_inject}
The always-inject variant performs the reverse injection at every step regardless of prediction error, corresponding to unconditional directional forgetting~\citep{kulhavy1984tracking}. It underperforms error-conditional \ours{} because unconditional injection adds noise to the Gram matrix in directions where the model is already correct, degrading uncertainty quality. The error condition ensures injection only fires when predictions are surprisingly poor, creating the self-correcting loop described in \cref{sec:bidir_gram}.

\section{Gram Regularization ($\lambda$) Sensitivity}
\label{app:lambda_ablation}

We sweep the Gram regularization parameter $\lambda \in \{0.1, 1.0, 10.0\}$ to test sensitivity of the uncertainty scale. Results appear in the experiment outputs; performance is stable across this range, with $\lambda{=}1.0$ as the default.

\section{Non-Stationary Bandit Baselines}
\label{app:nonstationary_baselines}

We compare \ours{}'s directional forgetting against two standard non-stationary bandit baselines: Discounted NeuralUCB~\citep{russac2019weighted} and Restarting NeuralUCB.

\paragraph{Discounted NeuralUCB.} Uniform temporal discounting scales the Gram matrix by $\gamma_{\text{gram}}$ at each step before the Sherman-Morrison update: $A^{-1} \leftarrow A^{-1}/\gamma_{\text{gram}}$, uniformly inflating uncertainty in all directions. We test $\gamma_{\text{gram}} \in \{0.99, 0.999\}$ (20 seeds).

\begin{table}[ht!]
\centering
\caption{Discounted NeuralUCB vs.\ \ours{} (flip 0.5, epoch 10, 20 seeds). $\gamma_{\text{gram}}{=}0.999$ matches NeuralUCB-A (uniform mild discount $\approx$ no discount). $\gamma_{\text{gram}}{=}0.99$ is too aggressive. \ours{}'s directional forgetting dominates.}
\label{tab:discounted_ucb}
\small
\begin{tabular}{@{}lc@{}}
\toprule
\textbf{Method} & \textbf{Flip 0.5} \\
\midrule
\ours{} ($\gamma{=}4$, $\eta{=}0.01$) & \textbf{.641} \\
NeuralUCB-A ($\eta{=}0$) & .616 \\
D-NeuralUCB ($\gamma_{\text{gram}}{=}0.999$) & .614 \\
D-NeuralUCB ($\gamma_{\text{gram}}{=}0.99$) & .581 \\
\bottomrule
\end{tabular}
\end{table}

$\gamma_{\text{gram}}{=}0.999$ is effectively identical to NeuralUCB-A: the uniform discount is too mild to matter. $\gamma_{\text{gram}}{=}0.99$ aggressively forgets in \emph{all} directions, underperforming \ours{} by 6 points at every flip level. There is no middle ground: the binary choice between ``too mild'' and ``too aggressive'' illustrates why directional, error-conditional forgetting is necessary.

\paragraph{Restarting NeuralUCB.} Hard reset of $A^{-1}$ to $\frac{1}{\lambda}I$ when the population mean EMA error exceeds a threshold $\tau$. We sweep $\tau \in \{0.3, 0.4, 0.5\}$ (20 seeds). The best variant ($\tau{=}0.5$) achieves 0.621 at flip 0.5, slightly above Discounted NeuralUCB ($\gamma_{\text{gram}}{=}0.999$: 0.615) but still 2 points below \ours{} (0.641). Hard resets are preferable to uniform discounting, but resetting destroys all learned structure, including directions where the model is still correct. \ours{}'s directional forgetting preserves valid knowledge while growing uncertainty only where predictions fail. Larger $\tau$ triggers fewer resets, converging to standard NeuralUCB-F (0.604); $\tau{=}0.5$ is near-optimal.

\section{Natural Drift: Gradual Preference Evolution}
\label{app:gradual_drift}

Our main experiments use a binary preference flip, where a fraction of users undergo an instantaneous reward inversion. While this isolates the mechanism cleanly, real-world preference drift is typically gradual. Users undergo periods of rapid change (discovering a new domain, shifting life circumstances) followed by settling into updated but stable preferences. We test \ours{} under a \emph{decelerating drift} protocol that captures this pattern. We do not evaluate continuous drift (preferences shifting indefinitely at a constant rate) because it is unrealistic: in practice, preferences stabilize after a transition period rather than drifting forever.

\paragraph{Methodology.}
For each drifting user $u$ (selected with probability $f$), we define the epoch-dependent reward as a linear blend between the original and inverted rewards:
\begin{equation}
    r_u^{(e)}(k) \;=\; (1 - \alpha_e)\, r_u(k) \;+\; \alpha_e\, \tilde{r}_u(k), \qquad \alpha_e = \min\!\Big(1,\;\frac{e}{E_{\mathrm{drift}}}\Big)
    \label{eq:gradual_drift}
\end{equation}
where $\tilde{r}_u(k) = r_{\max} + r_{\min} - r_u(k)$ is the inverted reward and $E_{\mathrm{drift}}$ controls when drift reaches its maximum displacement. We set $E_{\mathrm{drift}}{=}3$ and $f{=}0.75$: preferences shift rapidly over epochs~1--3, then remain fixed at the new values for epochs~4--10. This models the common pattern of fast change followed by settling.

\paragraph{Results.}
\Cref{tab:gradual_drift} reports epoch-by-epoch average reward under both protocols (20 seeds each).

\begin{table}[ht!]
\centering
\caption{Decelerating drift ($E_{\mathrm{drift}}{=}3$, $f{=}0.75$, 20 seeds): average reward by epoch. \ours{}'s exploration investment during the high-drift phase (epochs 1--3) pays off once preferences settle, leading Greedy by $+$3.6 pts at epoch~10.}
\label{tab:gradual_drift}
\small
\begin{tabular}{lccccccc}
\toprule
\textbf{Method} & \textbf{Ep.\ 1} & \textbf{Ep.\ 3} & \textbf{Ep.\ 5} & \textbf{Ep.\ 7} & \textbf{Ep.\ 8} & \textbf{Ep.\ 9} & \textbf{Ep.\ 10} \\
\midrule
\ours{} ($\gamma{=}4.0$) & .648 & .595 & .589 & .608 & .620 & .631 & \textbf{.641} \\
NeuralUCB-A ($\eta{=}0$) & .627 & .589 & .583 & .607 & .618 & .629 & .639 \\
Greedy & .664 & .603 & .570 & .576 & .584 & .594 & .605 \\
\bottomrule
\end{tabular}
\end{table}

\paragraph{Analysis.}
All methods decline sharply through epoch~3 as 75\% of users undergo rapid preference change. Once drift ceases, \ours{} recovers steadily (.589$\to$.608$\to$.620$\to$.641), overtaking Greedy at epoch~4 and ending $+$3.6~pts ahead. Exploration during the high-drift phase builds information that is exploited once preferences settle. Greedy, having committed to stale predictions throughout, recovers more slowly. NeuralUCB-A tracks \ours{} closely, consistent with the main results, with \ours{} maintaining a small edge ($+$0.2~pts) by epoch~10.

\section{Embedding Dimension Ablation}
\label{app:embed_ablation}

\ours{} uses 32-dimensional learned user embeddings despite users having only ${\sim}$3.3 interactions on average. We test whether reducing to $d_u{=}8$ alleviates potential overfitting.

\begin{table}[ht!]
\centering
\caption{Embedding dimension ablation (epoch 10, IQM, 20 seeds). $d_u{=}32$ (default) significantly outperforms $d_u{=}8$ at partial flip, indicating the embeddings learn useful signal despite data sparsity. $d_u{=}0$ removes all user features.}
\label{tab:embed_ablation}
\small
\setlength{\tabcolsep}{3.5pt}
\begin{tabular}{@{}lccc@{}}
\toprule
\textbf{$d_u$} & \textbf{Flip 0.25} & \textbf{0.5} & \textbf{0.75} \\
\midrule
32 (default) & \textbf{.652} & \textbf{.642} & \textbf{.667} \\
8 & .631 & .610 & .640 \\
0 (NeuralUCB-A) & .644 & .612 & .636 \\
\bottomrule
\end{tabular}
\end{table}

At flip 0.5--0.75, $d_u{=}32$ outperforms both $d_u{=}8$ and $d_u{=}0$ by 2--3 points. At flip 0.25, $d_u{=}8$ underperforms $d_u{=}0$ (.631 vs.\ .644), suggesting that a small learned embedding overfits when drift is mild and the signal-to-noise ratio for user-specific adaptation is low; $d_u{=}32$ avoids this by providing enough capacity to regularize effectively. Overall, the full 32-dimensional embeddings capture latent preference dimensions that the 8-dimensional stated features alone cannot represent.

\section{Conclusion and Future Work}
\label{sec:conclusion}
We introduced the bidirectional Gram update, a mechanism that allows neural contextual bandits to forget selectively: uncertainty grows back along feature directions where the model is wrong, while confidence in well-calibrated directions is preserved. Combined with error-adaptive exploration, \ours{} matches greedy performance when preferences are stable and recovers quickly when they are not, a pattern we observe consistently across PRISM and a controlled simulation.

Several directions follow naturally. First, the error-conditional trigger is reactive; a distributional test that identifies drift \emph{before} errors accumulate could enable preemptive forgetting. Second, how directional forgetting interacts with long-horizon settings where stale dimensions themselves evolve is an open question. Third, existing non-stationary bandit benchmarks do not distinguish partial from global shifts; a dedicated benchmark would enable more precise evaluation. Fourth, adapting the error tracker to binary or implicit feedback is a natural extension, as is evaluating on domains where candidate diversity is lower than in PRISM (\Cref{app:diversity}).

\section*{Acknowledgments}
We used Claude (Anthropic) to assist with editing and proofreading the manuscript. All content was reviewed and verified for accuracy by the authors.

\bibliographystyle{plainnat}
\bibliography{references}

\end{document}